# MathVerse: Does Your Multi-modal LLM Truly See the Diagrams in Visual Math Problems?

**Renrui Zhang**[*‡1,2], **Dongzhi Jiang**[*1], **Yichi Zhang**[*2], **Haokun Lin**[2], **Ziyu Guo**[2], **Pengshuo Qiu**[2]
**Aojun Zhou**[1], **Pan Lu**[3], **Kai-Wei Chang**[3], **Peng Gao**[†2], **Hongsheng Li**[†1]

[1]CUHK MMLab [2]Shanghai Artificial Intelligence Laboratory
[3]University of California, Los Angeles

{zhangrenrui, dzjiang, ziyuguo}@link.cuhk.edu.hk
lupantech@gmail.com, gaopeng@pjlab.org.cn, hsli@ee.cuhk.edu.hk

## Abstract

The remarkable progress of Multi-modal Large Language Models (MLLMs) has garnered unparalleled attention, due to their superior performance in visual contexts. However, their capabilities in visual math problem-solving remain insufficiently evaluated and understood. We investigate current benchmarks to incorporate excessive visual content within textual questions, which potentially assist MLLMs in deducing answers without truly interpreting the input diagrams. To this end, we introduce **MathVerse**, an all-around visual math benchmark designed for an equitable and in-depth evaluation of MLLMs. We meticulously collect 2,612 high-quality, multi-subject math problems with diagrams from publicly available sources. Each problem is then transformed by human annotators into six distinct versions, each offering varying degrees of information content in multi-modality, contributing to **15K** test samples in total. This approach allows MathVerse to comprehensively assess *whether and how much MLLMs can truly understand the visual diagrams for mathematical reasoning.* In addition, we propose a Chain-of-Thought (CoT) evaluation strategy for a fine-grained assessment of the output answers. Rather than naively judging True or False, we employ GPT-4(V) to adaptively extract crucial reasoning steps, and then score each step with detailed error analysis, which can reveal the intermediate CoT reasoning quality by MLLMs. With MathVerse, we unveil that, most existing MLLMs struggle to understand math diagrams, relying heavily on textual questions. Surprisingly, some of them even achieve 5%+ higher accuracy without the visual input, e.g., Qwen-VL-Max and InternLM-XComposer2. In contrast, GPT-4V and ShareGPT4V demonstrate relatively better comprehension of the visual content for mathematical reasoning. We hope MathVerse may provide unique insights to guide the future development of MLLMs. Project page: https://mathverse-cuhk.github.io.

## 1 Introduction

With the substantial advances of big data and computational power, Large Language Models (LLMs) [4, 28, 55, 56, 13], such as ChatGPT [45] and GPT-4 [46], have emerged as a central point of interest in both industry and academia. To broaden their applicability across diverse contexts, Multi-modal Large Language Models (MLLMs) [66, 20, 52, 11, 61, 70] have recently become

[1]* Equal contribution ‡ Project lead † Corresponding author

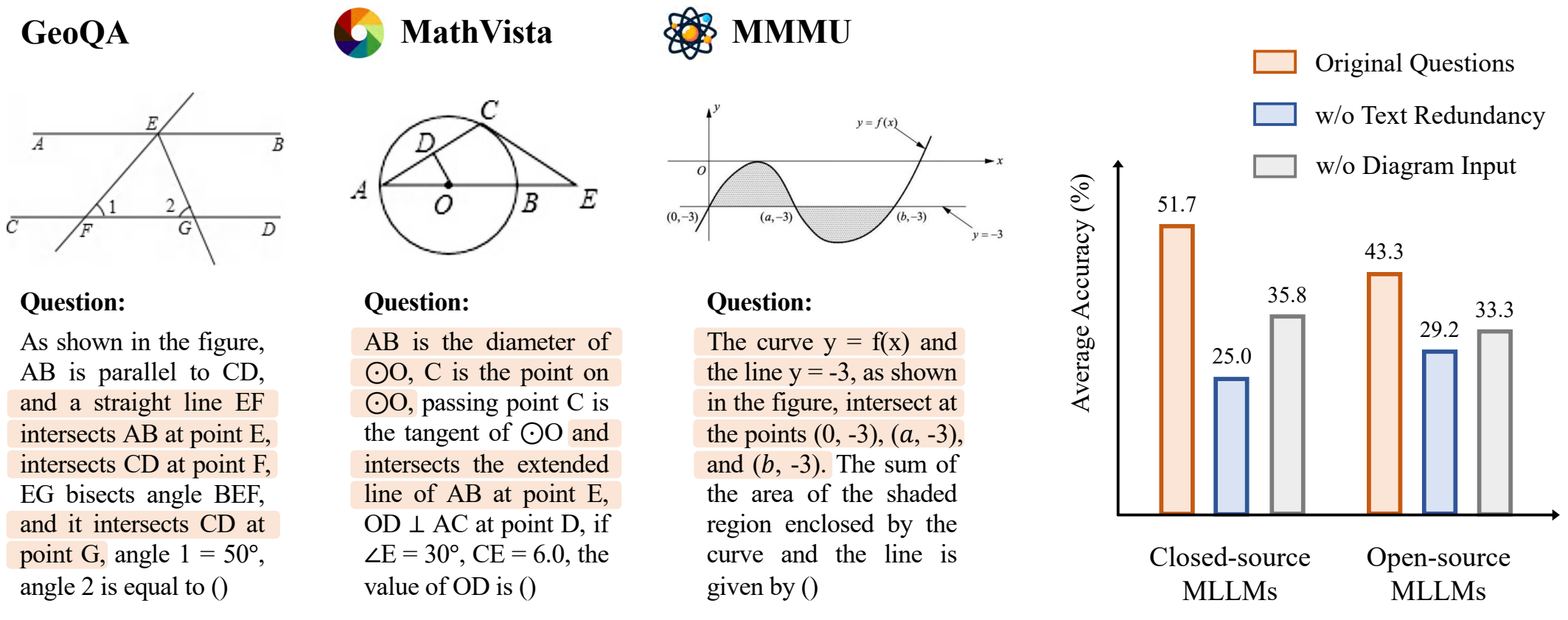

(a) **Text Redundancy** within Existing Benchmarks (b) Ablation Study

Figure 1: **(a)** We showcase three examples of **Text Redundancy** (highlighted in red) within existing visual math benchmarks [9, 41, 63]. **(b)** We report an ablation study by respectively removing the redundant texts and input diagrams on 120 randomly selected problems, for closed-sourced [47, 22, 3] and open-sourced [21, 38, 16] MLLMs.

a fast-evolving track, exemplified by the latest GPT-4V [47], Gemini [22], and the open-source LLaVA [39, 34, 32, 30] and SPHINX [36, 21]. Concurrently, a diverse array of evaluation benchmarks [17, 40, 33, 18, 53] are curated to assess their visual comprehension performance across different domains. Notably, the capability to solve mathematical problems involving diagrams serves as a critical measure, offering insights into the multi-modal logical thinking prowess of MLLMs. This task demands MLLMs to accurately decode the visual elements within input diagrams (characters and figures), and correlate them with the condition specified by textual questions for mathematical reasoning. Previous efforts [42, 51], e.g., GeoQA [9, 5] and UniGeo [7], concentrate on the challenging geometric problems, while the recent MathVista [41] and MMMU [63] expand the scope to encompass broader disciplines, including functions, charts, and scientific problems.

However, through our comprehensive observation and analysis, we identify three primary issues in current mathematical benchmarks for evaluating MLLMs:

i. **Do MLLMs truly see the math diagrams in evaluation?** This is the most fundamental question concerning the accurate assessment of visual math problem-solving. In Figure 1 (a), we showcase three examples from current benchmarks. We observe their texts contain too much duplicate information (highlighted in red) that is also depicted in the diagram. This redundancy might inadvertently provide MLLMs with a shortcut to resolve the problem by mostly reading the text, rather than interpreting the diagram. Our hypothesis gains support from the experiment in Figure 1 (b). For 40 randomly sampled problems from each benchmark, we remove such redundant texts from the question, challenging MLLMs to capture the corresponding information exclusively from visual inputs. The results reveal a significant drop in accuracy among most MLLMs (the blue column), even falling below the scores without taking diagrams as input (the grey column). This outcome suggests that ***MLLMs primarily depend on textual cues rather than the visual diagrams themselves to solve these problems in evaluation.*** Given this, we demonstrate that current visual math benchmarks might not be comprehensive enough to assess the genuine multi-modal mathematical reasoning capabilities of MLLMs.

ii. **Is it equitable to assess solely by the final answer?** Most existing multi-modal benchmarks directly compare model outputs with ground truths to derive a binary evaluation result. While this approach may suffice for general visual contexts, it falls short in math problems that require intricate step-by-step reasoning. In Figure 2, we examine three model outputs. Although they all arrive at incorrect answers in the end, they demonstrate varying levels of precision in the intermediate reasoning processes. Merely categorizing these outputs as 'Incorrect' fails to capture the nuanced differences in the reasoning quality of MLLMs.

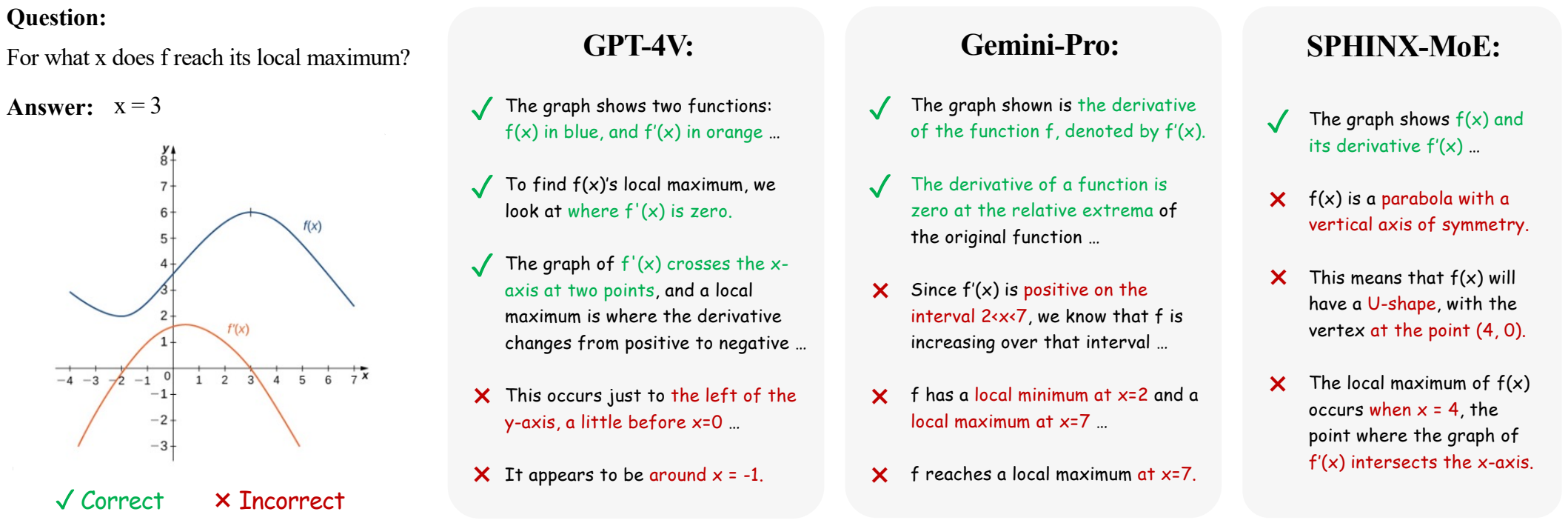

Figure 2: **Comparison of Visual Mathematical Reasoning by Three MLLMs.** Despite the incorrect final answer, GPT-4V [47], Gemini-Pro [22], and SPHINX-MoE [21] exhibit different levels of quality in the intermediate reasoning process.

iii. **Do they specialize in mathematical reasoning evaluation?** GeoQA, UniGeo, and other previous works narrowly target specific aspects of plane geometry. This limits the evaluation of broader mathematical capabilities, e.g., functions and solid geometry. Instead, MathVista expands its scope by including a wide array of peripheral tasks (19 out of 28), encompassing natural images, statistic plots, and charts, which do not directly evaluate professional math skills. Furthermore, the math problems in MMMU are of college-level complexity with extensive domain-specific knowledge, potentially hindering MLLMs from fully demonstrating their reasoning capacity.

Therefore, in light of the issues discussed, we present **MathVerse**, a holistic and specialized visual math benchmark crafted to evaluate the multi-modal mathematical reasoning skills of MLLMs. This benchmark encompasses a meticulously collected dataset of 2,612 visual math problems, with 1,236 newly acquired from public question repositories and 1,376 selected from existing benchmarks, ensuring a diverse range of challenges. To specialize in mathematical reasoning, MathVerse spans three primary areas: plane geometry, solid geometry, and functions. Each problem has been rigorously reviewed by expert annotators and classified into twelve detailed categories, emphasizing different fine-grained problem-solving capabilities. Notably, MathVerse distinguishes itself by introducing two novel strategies for evaluating MLLMs.

First, we investigate the influence of textual redundancy and validate whether MLLMs can interpret the diagrams for mathematical reasoning. As illustrated in Figure 3 (Left), we categorize the textual content within the questions into three different types: *Descriptive Information*, *Implicit Property*, and *Essential Condition*. These categories, arranged in ascending order of significance for problem-solving, correspond to information directly observable from the diagram, implicit spatial properties that demand advanced visual perception, and specific measurements crucial for computing the solution, respectively. Based on this problem formulation, expert annotators progressively remove the textual information from the questions in MathVerse, while incrementally incorporating elements into the visual diagrams to ensure problems are adequately defined. As shown in Figure 3 (Right), this process results in six unique versions of each problem characterized by a reduction in textual content and an enhancement in visual elements, creating a total of 15K test samples. These delicately curated problems can indicate the various multi-modal capabilities of MLLMs, such as geometric element understanding, function curve perception, and numerical value recognition, which thoroughly unveils whether and how much they comprehend the visual diagram for mathematical reasoning.

Second, to rigorously assess the visual Chain-of-Thought (CoT) capabilities [58], we propose a **CoT Evaluation strategy** for the step-by-step reasoning assessment of MLLMs. For each model's output, we leverage GPT-4 to first extract several crucial steps exclusively from the solving process, deliberately omitting the input of the question and answer. This approach aims to mitigate the bias towards GPT-4's inherent question-answering propensities. Then, the corresponding question, diagram, and ground-truth answer are fed into GPT-4 to evaluate each identified critical step, and provide detailed error analysis. Finally, the overall score is obtained by considering every single step within reasoning. Note that, we do not pre-define a ground-truth key-step template, since each math problem may encompass a variety of solution pathways, and different MLLMs tend to exhibit variable reasoning lengths. With CoT scoring, MathVerse showcases a fine-grained evaluation of the intermediate logical deduction of MLLMs, demonstrating visual mathematical CoT capabilities.

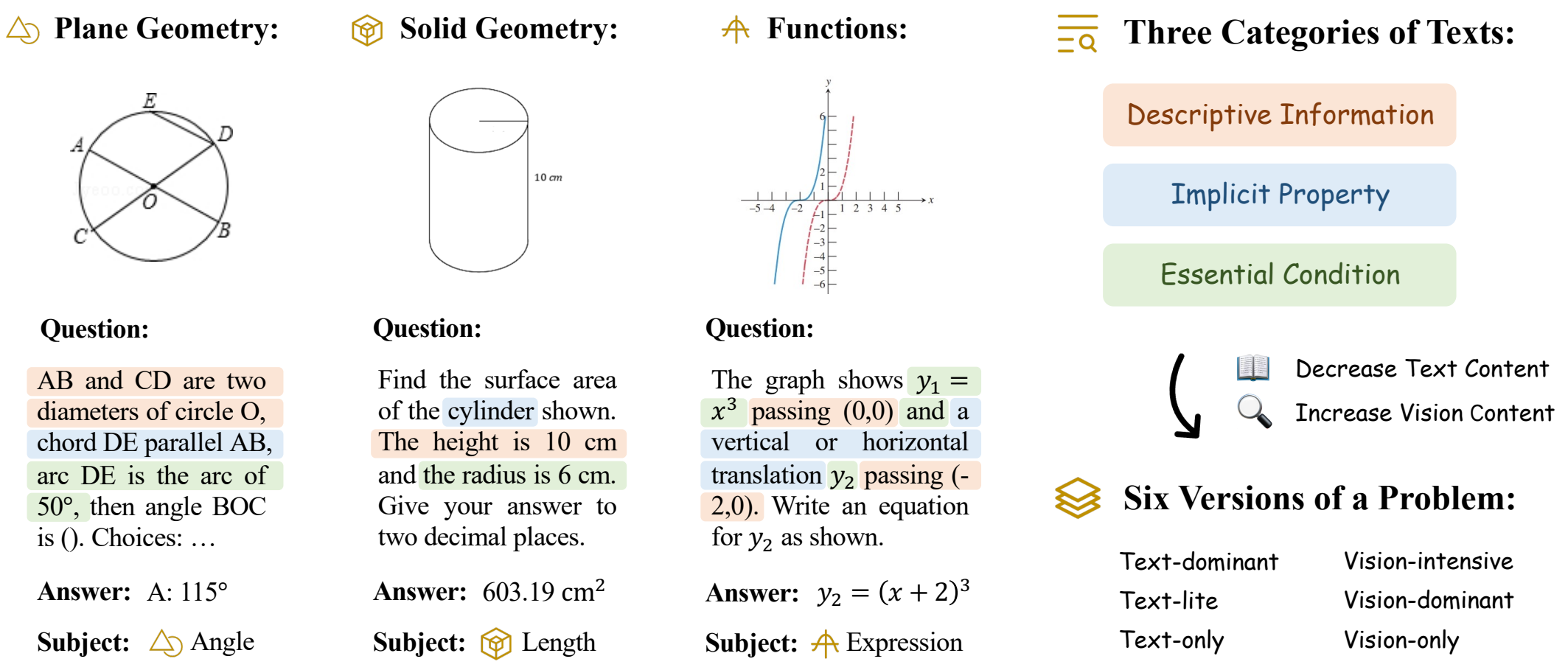

Figure 3: **Three Categories of Question Texts in MathVerse.** According to the significance for problem-solving, we categorize the question texts into three categories, and transform each problem into six versions for evaluation, with varying content in multi-modality. We present three examples in MathVerse for illustration.

We conduct extensive experiments on MathVerse with popular closed-source [47, 3, 22] and open-source [37, 38, 16, 21] MLLMs. Comparing different problem versions, we unveil that, most existing MLLMs struggle to understand math diagrams, relying heavily on textual questions. Therein, GPT-4V [47] achieves the best overall performance across different problem versions and subjects. Surprisingly, some of the MLLMs even attain much higher results without the diagram input, e.g., +5.1% for Qwen-VL-Max [3] and +5.6% for InternLM-XComposer2 [16]. With the fine-grained error analysis produced by our CoT evaluation strategy, we demonstrate such results are due to their deficient visual encoding capacity for mathematical diagrams, which instead acts as a distraction for problem-solving. In contrast, GPT-4V and ShareGPT4V [12] demonstrate relatively better comprehension of the visual content for mathematical reasoning. Our experimental results suggest that inadequate mathematical visual interpretation capabilities represent the most significant impediment for MLLMs in addressing multi-modal math problems, indicating substantial potential for advancement.

The contributions of this paper are summarized as follows:

- We investigate primary issues within existing benchmarks and introduce MathVerse, an all-around multi-modal benchmark evaluating the visual mathematical reasoning of MLLMs. The meticulously curated dataset contains 20K test problems with diagrams for a comprehensive assessment.
- By modifying problems with varying information content in multi-modality, we explore whether and how much MLLMs can understand the visual diagrams for mathematical reasoning, rather than relying on question texts.
- We propose a CoT evaluation strategy with GPT-4 to extract and assess each key step in the reasoning process of MLLMs, which provides a detailed error analysis and fine-grained evaluation of their multi-modal mathematical CoT capabilities.

# 2 MathVerse

In Section 2.1, we first present an overview of the curated visual math dataset in MathVerse. Then, in Section 2.2, we introduce our data formulation approach for investigating the visual mathematical comprehension of Multi-modal Large Language Models (MLLMs). Finally, in Section 2.3, we elaborate on the methodology of our proposed Chain-of-Thought (CoT) evaluation strategy.

Table 1: **Key Statistics of MATHVERSE.**

| Statistic | Number |
|---|---|
| Total questions | 2,612 |
| - Multiple-choice questions | 1,631 (62.4%) |
| - Free-form questions | 981 (37.6%) |
| - **Newly collected questions** | **1,236 (47.3%)** |
| - Existing-dataset questions | 1,376 (52.7%) |
| - **Questions with explanations** | **1,236 (47.3%)** |
| **Total test samples** | **15,672** |
| - **Newly annotated samples** | **10,448 (66.7%)** |
| - Samples of each version | 2,612 (16.7%) |
| Number of unique images | 2,420 (92.6%) |
| Number of unique questions | 2,573 (98.5%) |
| Number of unique answers | 847 (32.4%) |
| Maximum question length | 203 |
| Maximum answer length | 17 |
| Average question length | 35.7 |
| Average answer length | 1.4 |

Figure 4: **Subject Distribution of MATHVERSE.** Solid G: Solid Geometry, Plane G: Plane Geometry.

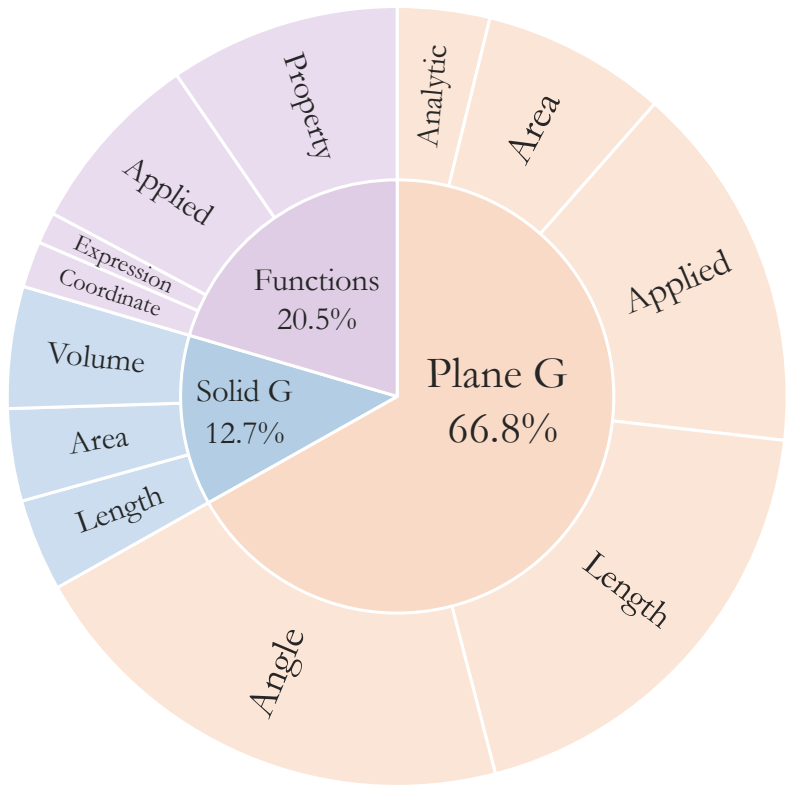

## 2.1 Visual Math Dataset

To thoroughly assess visual mathematical proficiency, we compile a comprehensive problem set covering a broad spectrum of math subjects, diagram patterns, and specialized knowledge domains. This widespread collection for MATHVERSE aims to pose diverse challenges to MLLMs, ensuring a robust evaluation of their capabilities in visual contexts.

**Data Composition and Categorization.** MATHVERSE comprises a total of 2,612 visual math problems, which contribute to the final created 15K test samples. Detailed statistics for data composition are presented in Table 1. This meticulously collected dataset covers three fundamental math subjects, i.e., plane geometry (1,746), solid geometry (332), and functions (534), where the latter two are all composed of newly collected problems. The choice of these three subjects is not only due to their rigorous demands on multi-modal reasoning, but also for two other considerations. For one thing, as we specialize MATHVERSE in mathematical problem-solving, other peripheral tasks in MathVista [41] are not included, e.g., statistical reasoning, table question-answering, and puzzle tests. For another, we expect the evaluation to fully display the reasoning capabilities of MLLMs with moderate-level mathematical knowledge. This avoids limiting their performance with overly complex domain-specific theorems or prior commonsense knowledge. Therefore, we deliberately focus the collected problems on the high school level, excluding advanced college-level disciplines like calculus and graph theory featured in MMMU [63]. Furthermore, expert annotators subdivide the problems into twelve fine-grained categories, as depicted in Figure 4, showcasing various dimensions of visual mathematical skills.

**Data Collection and Review Process.** Our collection procedure for high-quality visual math problems involves a rigorous selection from both pre-existing datasets and public question repositories. In the domain of plane geometry, we initially select 750 problems from GeoQA [9], 119 from GEOS [51], and 507 from Geometry3K [42], based on their original data quality and distribution. We exclude questions that are extremely simple or excessively complex, as well as those that appear dubious or lack necessary conditions. To enhance the diversity of question types and diagram styles, we further enrich our dataset with additional 370 plane geometry problems by manually collecting from other sources[1,2,3]. Given the scarcity of solid geometry and function-related problems in existing benchmarks, we purposefully gather these two types of problems (332 and 534, respectively) from new sources[1,2,3] to address this gap. Problems that include multiple diagrams or require visual illustrations within solutions are excluded, considering the current limitations of MLLMs in resolving such information. Note that, all the newly collected problems (1,236) accompany detailed explanations. After the preliminary collection, we undertake a comprehensive review to verify the accuracy of the answers, ensure consistency between questions and diagrams, and confirm the relevance of each problem to the defined twelve categories. This meticulous review guarantees the dataset's quality and precision.

[1]https://homework.study.com
[2]https://www.ixl.com/math
[3]https://mathspace.co/us

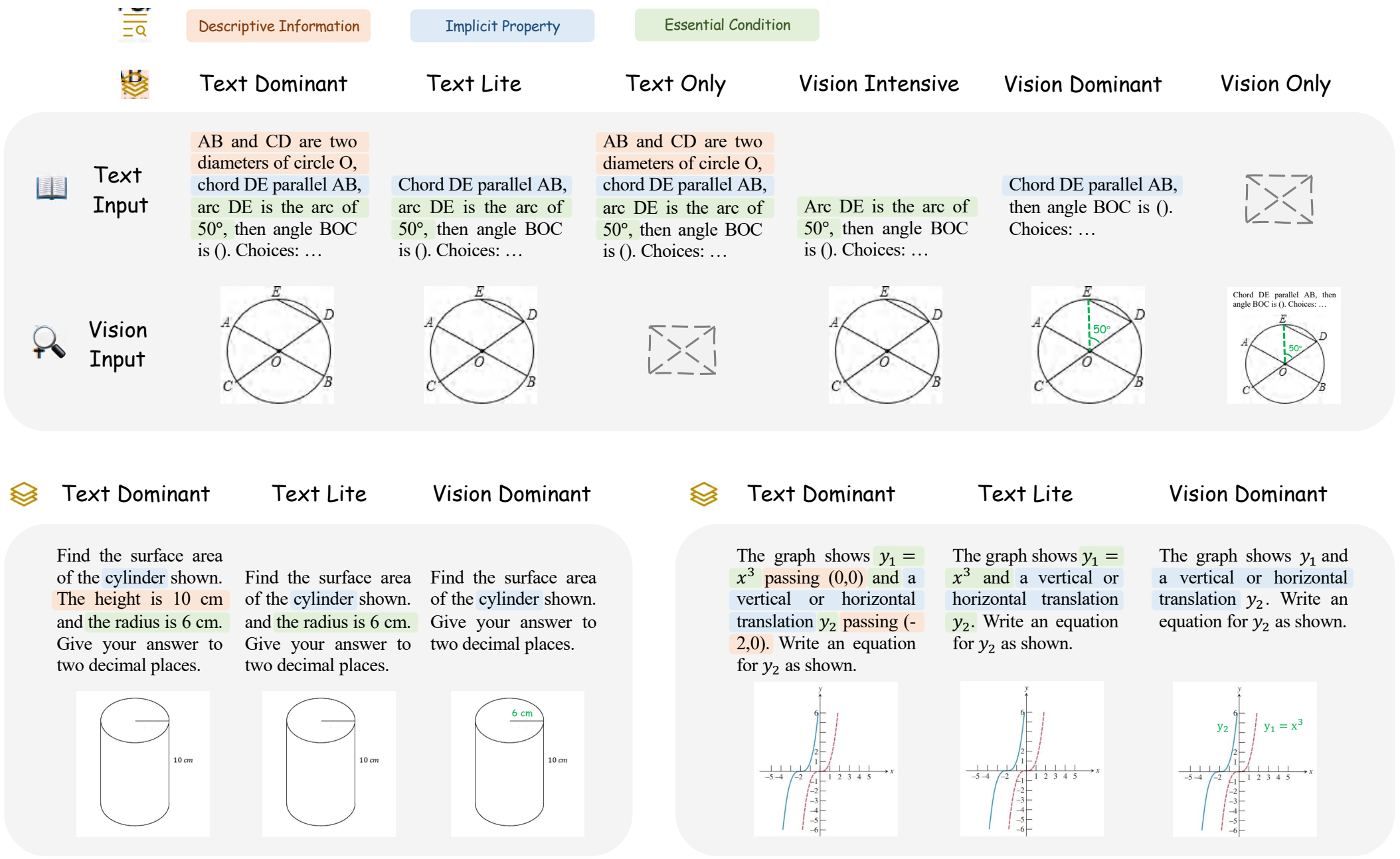

Figure 5: **Six Versions of Each Problem in MathVerse.** Expert annotators meticulously transform each visual math problem within MathVerse into six versions. They contain different vision-language content for a holistic visual mathematical evaluation.

## 2.2 Whether MLLMs Truly See the Diagrams?

In this section, we detail our data formulation approach to transform each problem in MathVerse into six different versions with varying information content in multi-modality. In this way, we explore the visual diagram understanding capabilities of MLLMs for mathematical reasoning.

**Three Types of Textual Information.** Considering the textual redundancy in original math problems, we first define three distinct categories for the textual information within the questions, as illustrated in Figure 3 and the following:

- ***Descriptive Information*** (**DI**) refers to the directly observable and clearly portrayed content in the diagram. It depicts the basic figure composition, spatial arrangement, and annotated entities, such as *the presence of geometric shapes or intersection points of functions*. These sentences normally help establish the context and frame the problem to orient the solver. Nevertheless, such information is repetitive to the visual components present in the diagram, thus regarded as redundant information for problem-solving. More importantly, it may assist MLLMs in bypassing the process of diagram interpretation, thereby undermining the assessment for visual mathematical reasoning, as evidenced in Figure 1.

- ***Implicit Property*** (**IP**) involves the information that requires a higher level of visual perception but less mathematical knowledge to discern from the diagram. It signifies strong visual conditions for problem-solving, such as *the parallelism and perpendicularity between lines, the similarity and congruence among triangles, and the category and periodicity of functions*. They can, in theory, be fully extracted from the diagrams alone, giving adequate capability for visual recognition and comprehension of MLLMs.

- ***Essential Condition*** (**EC**) denotes the specific numerical or algebraic measurements, which are indispensable conditions to derive the solution and cannot be derived from the visual diagram. This category encompasses precise values of angles, lengths, and function expressions, such as *an angle being 45 degrees, the length of BC being 6 units, and the functional equation* $f(x) = x^2 + 3$. Without these details in textual information, solving the visual math problem would be impossible.

**Creating Six Versions of Each Problem.** Based on the three categories, expert annotators systematically remove different textual information within questions, and incrementally incorporate the critical elements into diagrams. This approach can progressively reduce textual redundancy and information content, thereby increasingly compelling MLLMs to capture mathematical conditions from the visual input. As compared in Figure 5, we generate six versions of each problem in MathVerse, obtaining 15,672 test instances. With this curated problem set, we can provide a holistic evaluation of the genuine visual comprehension of MLLMs, and whether it can facilitate multi-modal mathematical reasoning. The details of each problem version are as follows:

- **Text-dominant Version** retains the entire textual content, including the three types of textual information and the question statement. If the original problem contains limited *Descriptive Information*, we manually add it within the textual content. This version may induce MLLMs to regard the text as the primary source of information, treating the diagram more as a supplementary visual aid. This serves as the baseline point for evaluation.

  📖 Text: **DI** + **IP** + **EC** + Question    🔍 Vision: Diagram    (1)

- **Text-lite Version** diminishes the *Descriptive Information* from the Text-dominant version, assuming this information can be observed from the diagram. This creates a condensed question without redundancy, forcing MLLMs to interpret the diagram for basic information.

  📖 Text: **IP** + **EC** + Question    🔍 Vision: Diagram    (2)

- **Text-only Version** directly discards the diagram input from the Text-dominant version. Comparing this to the Text-lite version helps identify where MLLMs mainly obtain the contextual visual information for problem-solving, the *Descriptive Information* or diagram.

  📖 Text: **DI** + **IP** + **EC** + Question    🔍 Vision: $\varnothing$    (3)

- **Vision-intensive Version** further removes the *Implicit Property* from the Text-lite version. Without the strong visual condition in texts, MLLMs are challenged to intensively leverage their visual interpretation skills to gather sufficient cues for mathematical reasoning. The outcome demonstrates their proficiency in understanding mathematical relationships visually.

  📖 Text: **EC** + Question    🔍 Vision: Diagram    (4)

- **Vision-dominant Version**, building upon the Text-lite version, excludes the *Essential Condition* from texts, instead annotating these measurements visually in diagrams. The textual content is narrowed down to *Implicit Property* and question statements. It demands MLLMs to recognize the *Essential Condition* exclusively from diagrams, and accurately correlate it with corresponding visual elements for problem-solving.

  📖 Text: **IP** + Question    🔍 Vision: Diagram + **EC**    (5)

- **Vision-only Version** strips away the entire textual input, conveying the problem solely through the diagram. We render the remaining textual content in the Vision-dominant version onto the diagram, and minimize the text to an almost negligible prompt. This compels MLLMs to extract figures and recognize questions both from the visual information, standing as an ultimate test for their visual reasoning capabilities in mathematics.

  📖 Text: $\varnothing$    🔍 Vision: Diagram + **EC** + **IP** + Question    (6)

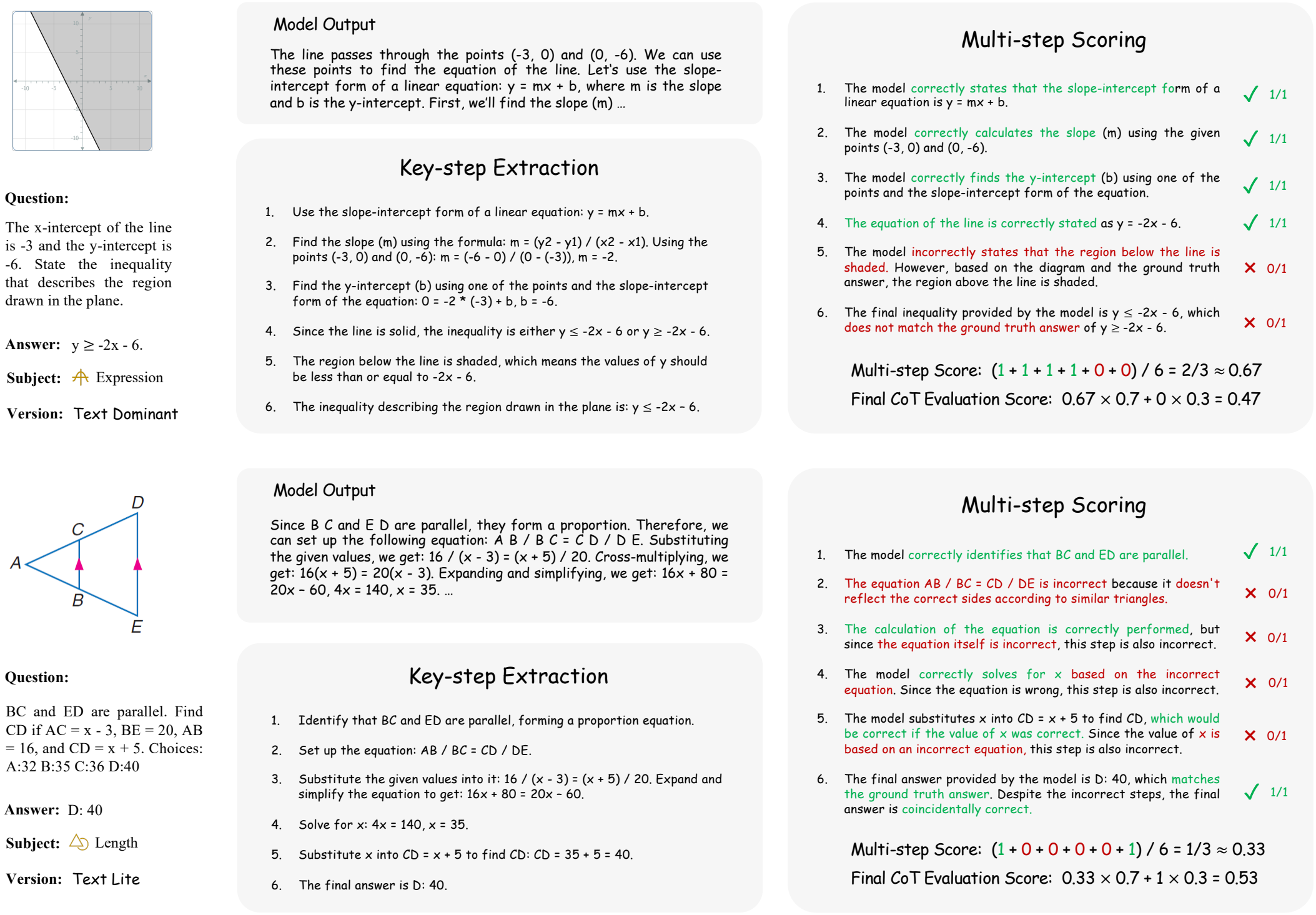

Figure 6: **Examples of the CoT Evaluation Strategy for MATHVERSE.** We present two outputs from Qwen-VL-Max [3] with our CoT evaluation strategy, which assesses the fine-grained reasoning capabilities with a detailed explanation for error analysis.

### 2.3 CoT Evaluation Strategy

Compared to visual question-answering in general scenarios, the solving process of MLLMs for mathematical problems requires nuanced, step-by-step CoT reasoning. Considering two cases in Figure 6, one arrives at the correct solution albeit through incorrect intermediary steps, while the other demonstrates the opposite phenomenon. Therefore, the binary 'Correct' or 'Incorrect' evaluative approach of existing benchmarks is inadequate to examine the depth and precision of the multi-step reasoning process. To this end, we propose a CoT evaluation strategy to thoroughly assess their mathematical CoT skills in visual contexts, involving two prompting phases with GPT-4(V) [47, 46].

**Key-step Extraction.** Given the output of an MLLM, we first employ GPT-4, the language-only version, to extract $N$ pivotal steps within the reasoning sequence, denoted as $[s_1, s_2, \dots, s_N]$, including the final answer $s_A$. Such key steps include significant computational outcomes, the identification of visual components, and critical immediate inferences. Note that, we only prompt GPT-4 with the MLLM's output, deliberately omitting the original questions, diagrams, and ground-truth answers. This approach aims to mitigate the inherent bias of GPT-4 itself towards problem-solving and visual diagram interpretation, thereby concentrating solely on the logical coherence of the model output. In addition, we do not pre-define a ground-truth key-step template for each problem, but perform the extraction adaptively for the unique output of every MLLM. Since the problem potentially encompasses diverse possible solution pathways, and different MLLMs exhibit varying reasoning lengths and styles, the rigid template would harm the CoT evaluation accuracy.

**Multi-step Scoring.** After the extraction phase, we utilize GPT-4V, the multi-modal version, to evaluate each critical step and culminate a comprehensive score. We feed the extracted key steps, the original questions, diagrams, and ground-truth answers all into GPT-4V, contributing to a holistic assessment, e.g., numerical computations, logical deductions, and visual interpretations. Therein, we observe that GPT-4V occasionally struggles with accurately recognizing elements within functional diagrams, leading to unstable evaluation for related problems. We thereby annotate additional information for function problems and together feed into GPT-4V, ensuring the quality of visual evaluation. Specifically, GPT-4V assesses each $N$ intermediate step with a binary score of '1' (correct) or '0' (incorrect), and derives the overall score by aggregating the correctness of the final answer. We formulate the scoring process as

$$\text{Score}_{\text{final}} = \alpha\Big(\frac{1}{N}\sum_{i=1}^{N}\text{Score}(s_i)\Big) + (1-\alpha)\text{Score}(s_A), \tag{7}$$

where $\alpha$ denotes a balancing factor between the intermediate steps and the final answer $s_A$. We set $\alpha$ as 0.7 by default to underscore the significance of CoT reasoning. As exemplified in Figure 6, besides the fine-grained scoring, the CoT evaluation can also provide a detailed error analysis of each step, which is valuable and instructive for the development of MLLMs in the field.

Table 2: **Mathematical Evaluation on Six Problem Versions in MathVerse's *testmini* Set.** We calculate the 'All' score without averaging the 'Text Only' version. 'CoT-E' or 'Acc' denotes whether to employ the proposed CoT evaluation strategy or not. The highest accuracy for closed-source and open-source MLLMs is marked in red and blue respectively.

| Model | All | | Text Dominant | | Text Lite | | Text Only | | Vision Intensive | | Vision Dominant | | Vision Only | |
|---|---|---|---|---|---|---|---|---|---|---|---|---|---|---|
| | CoT-E | Acc | CoT-E | Acc | CoT-E | Acc | CoT-E | Acc | CoT-E | Acc | CoT-E | Acc | CoT-E | Acc |
| *Baselines* | | | | | | | | | | | | | | |
| Random Chance | - | 12.4 | - | 12.4 | - | 12.4 | - | 12.4 | - | 12.4 | - | 12.4 | - | 12.4 |
| Human | - | 64.9 | - | 71.2 | - | 70.9 | - | 41.7 | - | 61.4 | - | 68.3 | - | 66.7 |
| *LLMs* | | | | | | | | | | | | | | |
| ChatGPT [48] | - | - | 51.3 | 33.3 | 38.5 | 18.9 | 51.3 | 33.3 | - | - | - | - | - | - |
| GPT-4 [46] | - | - | 63.4 | 46.5 | 40.7 | 20.7 | 63.4 | 46.5 | - | - | - | - | - | - |
| *Closed-source MLLMs* | | | | | | | | | | | | | | |
| Qwen-VL-Plus [3] | 21.3 | 11.8 | 26.0 | 15.7 | 21.2 | 11.1 | 25.2 | 14.5 | 18.5 | 9.0 | 19.1 | 13.0 | 21.8 | 10.0 |
| Gemini-Pro [22] | 35.3 | 23.5 | 39.8 | 26.3 | 34.7 | 23.5 | 44.5 | 27.3 | 32.0 | 23.0 | 36.8 | 22.3 | 33.3 | 22.2 |
| Qwen-VL-Max [3] | 37.2 | 25.3 | 42.8 | 30.7 | 37.7 | 26.1 | 47.9 | 28.9 | 33.6 | 24.1 | 35.9 | 24.1 | 35.9 | 21.4 |
| GPT-4V [47] | 54.4 | 39.4 | 63.1 | 54.7 | 56.6 | 41.4 | 60.3 | 48.7 | 51.4 | 34.9 | 50.8 | 34.4 | 50.3 | 31.6 |
| *Open-source MLLMs* | | | | | | | | | | | | | | |
| LLaMA-Adapter V2 [20] | 5.8 | 5.7 | 7.8 | 6.2 | 6.3 | 5.9 | 3.9 | 2.7 | 6.2 | 6.1 | 4.5 | 4.2 | 4.4 | 6.1 |
| ImageBind-LLM [24] | 10.0 | 9.2 | 13.2 | 11.4 | 11.6 | 11.3 | 12.9 | 11.7 | 9.8 | 8.9 | 11.8 | 11.2 | 3.5 | 3.4 |
| mPLUG-Owl2 [62] | 10.3 | 5.9 | 11.6 | 6.6 | 11.4 | 6.3 | 13.8 | 6.1 | 11.1 | 6.3 | 9.4 | 5.6 | 8.0 | 4.9 |
| MiniGPT-v2 [11] | 10.9 | 11.0 | 13.2 | 12.1 | 12.7 | 12.0 | 15.3 | 11.7 | 11.1 | 13.1 | 11.3 | 10.3 | 6.4 | 7.4 |
| LLaVA-1.5 [37] | 12.7 | 7.6 | 17.1 | 8.8 | 12.0 | 7.6 | 22.6 | 11.5 | 12.6 | 7.4 | 12.7 | 7.4 | 9.0 | 6.9 |
| SPHINX-Plus [21] | 14.0 | 12.2 | 16.3 | 13.9 | 12.8 | 11.6 | 15.8 | 14.9 | 12.9 | 11.6 | 14.7 | 13.5 | 13.2 | 10.4 |
| G-LLaVA [19] | 15.7 | 16.6 | 22.2 | 20.9 | 20.4 | 20.7 | 21.6 | 21.1 | 16.5 | 17.2 | 12.7 | 14.6 | 6.6 | 9.4 |
| LLaVA-NeXT [38] | 17.2 | 15.6 | 21.6 | 19.4 | 19.7 | 15.2 | 25.1 | 18.1 | 17.6 | 16.8 | 14.9 | 15.2 | 12.1 | 11.3 |
| ShareGPT4V [12] | 17.4 | 13.1 | 21.8 | 16.2 | 20.6 | 16.2 | 14.6 | 6.6 | 18.6 | 15.5 | 16.2 | 13.8 | 9.7 | 3.7 |
| SPHINX-MoE [21] | 22.8 | 15.0 | 33.3 | 22.2 | 21.9 | 16.4 | 40.7 | 18.3 | 21.1 | 14.8 | 19.6 | 12.6 | 18.3 | 9.1 |
| InternLM-XC2. [16] | 25.9 | 16.5 | 36.9 | 22.3 | 28.3 | 17.0 | 42.5 | 16.5 | 20.1 | 15.7 | 24.4 | 16.4 | 19.8 | 11.0 |

# 3 Experiments

In this section, we conduct a systematic evaluation of existing Multi-modal Large Language Models (MLLMs) on MathVerse. We first introduce the experimental setup in Section 3.1. Then, we detail the quantitative results in Section 3.2 and narrate the error analysis in Section 3.3.

## 3.1 Experimental Setup

**Division of the *testmini* Subset.** MathVerse encompasses a comprehensive collection of 2,612 visual math problems, alongside 15,672 corresponding test instances. To enable faster evaluation and model development validation, we extract a smaller subset termed *testmini* including 788 problems and 4,728 instances. In constructing *testmini*, we employ a random sampling strategy across different subfields, maintaining a sample size proportional to the overall dataset to preserve its statistical representativeness. The remaining test set features 1,824 problems and 10,944 samples will be utilized for standard evaluation and publicly released in the future. ***In the subsequent experiments, all quantitative results are assessed using the testmini subset of* MathVerse.**

**Evaluation Models.** We examine the performance of foundation models across three distinct categories on MathVerse: (a) *Large Language Models (LLMs)* as the text-only baseline, which only take textual questions as input, including ChatGPT [45] and GPT-4 [46], (b) *Closed-source*

Table 3: **Mathematical Evaluation on Different Subjects and Subfields in MathVerse's *testmini* Set.** We report the scores averaging five problem versions except for the 'Text Only' version, and employ the CoT evaluation strategy by default. Len: Length; Anal: Analytic; Apply: Applied; Vol: Volume; Coord: Coordinate; Prop: Property; Exp: Expression; Apply: Applied. The highest accuracy for closed-source and open-source MLLMs is marked in red and blue respectively.

| Model | All | Plane Geometry | | | | | | Solid Geometry | | | | Functions | | | | |
|---|---|---|---|---|---|---|---|---|---|---|---|---|---|---|---|---|
| | | All | Len | Area | Angle | Anal | Apply | All | Len | Area | Vol | All | Coord | Prop | Exp | Apply |
| *Closed-source MLLMs* | | | | | | | | | | | | | | | | |
| Qwen-VL-Plus [3] | 21.3 | 17.3 | 19.1 | 16.4 | 16.1 | 23.6 | 13.2 | 24.8 | 18.1 | 18.7 | 33.4 | 31.3 | 52.5 | 25.1 | 10.8 | 50.3 |
| Gemini-Pro [22] | 35.3 | 33.0 | 32.2 | 42.6 | 28.4 | 30.2 | 32.3 | 33.4 | 35.0 | 29.3 | 36.1 | 28.3 | 25.7 | 26.6 | 10.8 | 51.3 |
| Qwen-VL-Max [3] | 37.2 | 38.4 | 41.7 | 46.4 | 32.6 | 40.6 | 38.7 | 33.7 | 25.4 | 28.3 | 42.6 | 38.4 | 43.7 | 35.5 | 13.6 | 61.0 |
| GPT-4V [47] | 54.4 | 56.9 | 60.8 | 63.4 | 52.6 | 48.5 | 60.9 | 50.2 | 54.8 | 39.9 | 56.8 | 52.8 | 72.3 | 47.1 | 30.9 | 70.1 |
| *Open-source MLLMs* | | | | | | | | | | | | | | | | |
| LLaMA-Adapter V2 [20] | 5.8 | 5.9 | 4.0 | 5.9 | 6.6 | 13.4 | 3.3 | 4.6 | 5.3 | 3.1 | 5.7 | 6.2 | 6.7 | 6.1 | 4.5 | 7.9 |
| ImageBind-LLM [24] | 10.0 | 9.7 | 12.1 | 9.9 | 9.2 | 10.2 | 4.8 | 4.6 | 4.9 | 3.5 | 5.3 | 14.9 | 12.3 | 13.8 | 4.6 | 25.9 |
| mPLUG-Owl2 [62] | 10.3 | 7.7 | 8.2 | 6.0 | 5.7 | 12.4 | 10.6 | 11.0 | 9.2 | 6.7 | 15.7 | 17.4 | 22.8 | 18.6 | 5.3 | 22.2 |
| MiniGPT-v2 [11] | 10.9 | 11.6 | 10.0 | 9.8 | 14.3 | 9.1 | 11.8 | 1.7 | 2.2 | 1.6 | 0.5 | 11.2 | 4.2 | 15.7 | 4.0 | 21.1 |
| LLaVA-1.5 [37] | 12.7 | 11.8 | 13.1 | 15.1 | 9.7 | 9.4 | 13.2 | 10.6 | 12.1 | 8.7 | 11.6 | 14.8 | 18.8 | 12.7 | 9.5 | 23.7 |
| SPHINX-Plus [21] | 14.0 | 14.4 | 14.2 | 10.5 | 14.1 | 16.5 | 16.8 | 7.0 | 7.2 | 6.1 | 7.6 | 17.9 | 11.1 | 19.1 | 6.3 | 27.7 |
| G-LLaVA [19] | 15.7 | 20.2 | 17.3 | 13.6 | 26.5 | 5.9 | 23.1 | 5.0 | 10.3 | 4.4 | 3.1 | 9.2 | 9.1 | 9.1 | 1.3 | 15.5 |
| LLaVA-NeXT [38] | 17.2 | 15.9 | 14.8 | 13.1 | 16.3 | 17.7 | 17.8 | 19.6 | 33.3 | 11.7 | 12.6 | 23.1 | 24.5 | 23.4 | 8.0 | 33.1 |
| ShareGPT4V [12] | 17.4 | 16.9 | 16.2 | 17.9 | 16.9 | 12.2 | 21.1 | 15.0 | 13.6 | 10.9 | 19.7 | 20.2 | 19.9 | 22.2 | 8.4 | 25.8 |
| SPHINX-MoE [21] | 22.8 | 24.5 | 26.3 | 28.4 | 21.1 | 26.6 | 24.4 | 15.8 | 9.4 | 10.7 | 26.3 | 19.5 | 23.5 | 19.3 | 9.2 | 30.3 |
| InternLM-XC2. [16] | 25.9 | 26.2 | 27.1 | 29.7 | 20.6 | 18.5 | 22.2 | 20.1 | 34.5 | 14.1 | 25.2 | 23.7 | 24.4 | 24.9 | 10.6 | 36.3 |

*MLLMs*, represented by models like GPT-4V [47], Gemini-Pro [22], Qwen-VL-Max [3], and Qwen-VL-Plus, and (c) *Open-source MLLMs*, featuring models such as LLaVA-1.5 [37] (Vicuna-13B [13]), LLaVA-NeXT [38] (Vicuna-13B), SPHINX-MoE [21] (Mixtral-8×7B [28]), SPHINX-Plus (LLaMA2-13B [56]), InternLM-XComposer2 [16] (InternLM2-7B [54]), LLaMA-Adapter V2 [20] (LLaMA-7B [55]), ImageBind-LLM [24] (LLaMA-7B), MiniGPT-v2 [11] (LLaMA2-7B), mPLUG-Owl2 [62] (LLaMA-7B), G-LLaVA [19] (LLaMA2-7B), and ShareGPT-4V [12] (Vicuna-13B).

**Implementation Details.** All our experiments are conducted under a zero-shot setting, showcasing the generalization capacity of MLLMs for mathematical reasoning, without few-shot prompting or further fine-tuning. By default, we employ the Chain-of-Thought (CoT) prompting technique [58], which encourages MLLMs to perform complete reasoning steps for a fine-grained evaluation. A baseline representing random chance is established for comparison, for which we select one option at random for multiple-choice questions and utilize empty for free-form questions. In addition, we recruit ten qualified college students, and ask them to solve the problems in MathVerse independently, serving as a baseline for human performance. We conduct all experiments on NVIDIA A100 GPUs. As the text-only LLMs can only take text questions as input, we evaluate them with the first three problem versions, i.e., Text Dominant, Text Lite, and Text Only. For the 'w/o' results, we utilize the template in MathVista [41] to prompt GPT-4 [46] for answer extraction, and directly score the final answer without the intermediate reasoning process.

### 3.2 Experimental Analysis

To best investigate the visual mathematical reasoning capabilities, we report the evaluation results of different models on MathVerse for the six transformed problem versions in Table 2 and twelve detailed subjects in Table 3. We mainly analyze the performance by the proposed Chain-of-Though (CoT) evaluation, and derive the following observations.

**MLLMs Rely More on DI than Seeing Diagrams.** Comparing the Text-dominant and Text-only versions, with the elimination of visual input, most MLLMs even obtain an unexpected performance improvement, e.g., +5.1% for Qwen-VL-Max and +5.6% for InternLM-XComposer2. This suggests that the unsatisfactory visual encoding for mathematical diagrams instead severely harms the original problem-solving capacity of MLLMs. As exemplified in Figure 7, from the error analysis of our CoT evaluation strategy, we observe that Gemini-Pro can deduce the correct answer exclusively by the visual information within the *Descriptive Information*. Instead, the inaccurate visual perception of mathematical elements directly interferes with the outcome of problem-solving, turning correct answers into incorrect ones. In contrast, GPT-4V and ShareGPT-4V achieve better results in Text Dominant than in Text Only, indicating their relatively better visual encoding, which would not degrade the performance. However, they still encounter a larger performance drop by removing the redundant *Descriptive Information* than the diagram input, e.g., GPT-4V and ShareGPT-4V. This pattern demonstrates that they tend to capture more visual information for mathematical reasoning from the text content, instead of seeing the diagram itself.

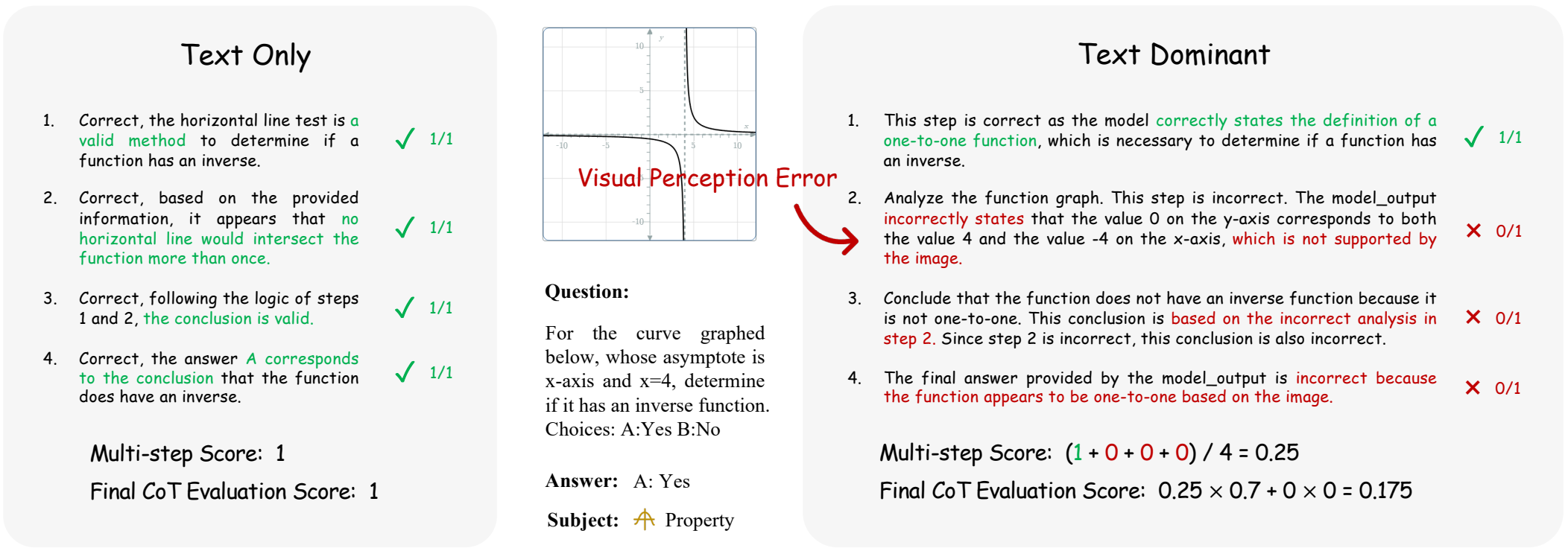

Figure 7: **A Typical Visual Perception Error by our CoT Evaluation Strategy.** The example is an output from Gemini-Pro [22], where the correct reasoning of the Text-only version is distracted by the visual perception error within the diagram.

**MLLMs are Moderately Effective at Perceiving IP.** By discarding the *Implicit Property* in question texts, a negligible decline in accuracy is noted from the Text-lite to Vision-intensive versions for most MLLMs. This is because the *Implicit Property* mainly encompasses the spatial layouts and geometric relationships, which demand minimal mathematical domain knowledge for interpretation. This outcome underscores the favorable visual perception skills of MLLMs for non-mathematical elements, which is not the primary obstacle hindering MLLMs in solving visual math problems.

**MLLMs are Challenged to interpret EC from Diagrams.** Incorporating the *Essential Condition* within diagrams challenges MLLMs to accurately identify and understand these conditions in vision modality for mathematical problem-solving. Evidence from the Vision-dominant results indicates a notable decline in the performance of most MLLMs compared to the Text-lite accuracy, such as -5.8% for GPT-4V and -3.9% for InterLM-XComposer2. This reveals their inaccurate identification of mathematical symbols and an insufficient grasp of domain-specific knowledge required to associate identified measurements with relevant concepts.

**MLLMs struggle to Solve Problems Entirely by Diagrams.** The scenario of Vision-only problems aligns more closely with real-world applications, where capturing an image is often more convenient than transcribing the problem into text. However, by rendering the whole question within the diagram, the mathematical problem-solving capacity of MLLMs is further diminished. This experiment unveils the great challenge for MLLMs to simultaneously understand mathematical conditions, questions, and figures from the visual input alone.

**Closed-source MLLMs are Better-performed.** From the performance in both tables, we observe a consistently better performance achieved by closed-source MLLMs than open-sourced ones. Despite the gap with humans, GPT-4V attains the leading position among MLLMs, showcasing superior mathematical capabilities over problem versions and subjects, especially the challenging subfields like 'Coord' and 'Prop' (the property and coordinate solving of function problems). InternLM-XComposer2 and SPHINX-MoE are the best-performing open-source MLLMs, while still lagging behind Gemini-Pro with a margin of 9.4% and 12.5% overall accuracy, respectively, suggesting large improvement space.

**LLMs Achieve Competitive Results to MLLMs.** Utilizing solely question texts as input, two LLMs, i.e., GPT-4 and ChatGPT, attain superior accuracy to most MLLMs in Text Dominant and Lite versions. Even in the absence of redundant *Descriptive Information* within Text-lite problems, GPT-4 outperforms InternLM-XComposer2 and SPHINX-MoE by substantial margins of 12.4% and 18.8%, respectively. These findings not only indicate the strong mathematical reasoning skills of LLMs, but further emphasize the deficiencies in diagram interpretation of existing MLLMs. Importantly, the performance of GPT-4 is only exceeded by GPT-4V, which demonstrates that a satisfactory diagram perception capability can enhance problem-solving for visual mathematics.

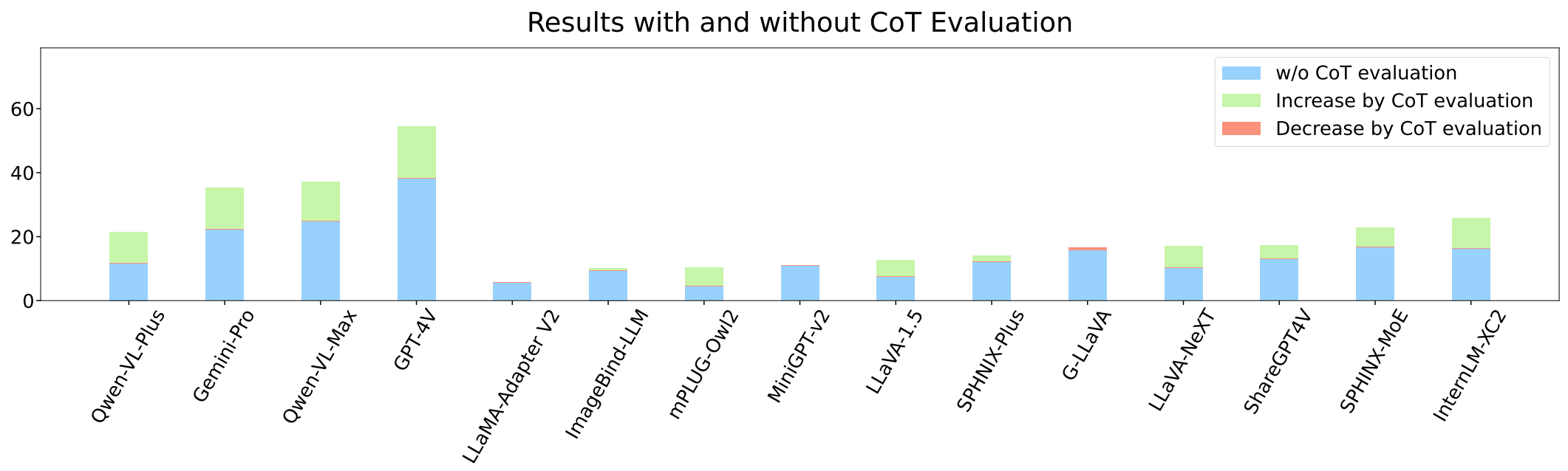

Figure 8: **Results with and without CoT Evaluation in MathVerse.** Referring to Table 2, we denote the 'w/o' results in blue pillars, and highlight the increase and decrease magnitude with 'CoT-E' by green and red colors, respectively.

**GPT-4(V) Beats Human in the Text-only Version.** Without the visual content provided in diagrams, human solvers often face challenges in deducing the correct answers due to the lack of sufficient information, e.g., 41.7% 'w/o' scores in Text-only problems. In contrast, GPT-4V and GPT-4 achieve the 'w/o' scores of 41.1% and 46.5%, respectively, which surpass the human performance. This comparison highlights their advanced reasoning capabilities in handling extreme scenarios, exhibiting more robustness for mathematical problem-solving given missing visual conditions.

**Mathematical Training Benefits the Performance.** In addition to foundational visual instruction-following datasets, both SPHINX-MoE and InternLM-XComposer2 extend their training regimes to include specialized mathematical problems that are either text-only or visual, such as MathQA [2], Geometry3K [43], and MathInstruct [64]. This approach of math-specific tuning contributes to their leading performance in MathVerse. Furthermore, G-LLaVA fine-tunes LLaVA-1.5 by a large-scale visual geometric dataset containing 170K enriched problems. This targeted refinement can improve several fields ('Len', 'Angle', and 'Apply') within the plane geometry subject. However, since G-LLaVA's fine-tuning data does not include problems of analytic geometry, solid geometry, and functions, it harms the related results of LLaVA-1.5 due to catastrophic forgetting, e.g., -3.5% in 'Anal', -5.6% in 'Solid Geometry', and -5.6% in 'Functions'. This phenomenon underscores the critical role of developing extensive, high-quality visual math data for effectively training MLLMs.

**Discrepancy Between 'CoT-E' and 'w/o' Scores.** As illustrated by Table 2, the 'CoT-E' scores for MLLMs, in most cases, are much higher than 'w/o' scores, e.g., +16.1% for GPT-4V and +9.6% for InternLM-XComposer2. This observation demonstrates that our proposed CoT evaluation strategy identifies numerous correct intermediate reasoning steps, despite the final incorrect answer, highlighting the effectiveness of fine-grained assessment. In Figure 8, we present the statistics of variance between 'CoT-E' and 'w/o' scores within different MLLMs. Although GPT-4V attains top-tier performance, it exhibits a pronounced gap of 16.1% concerning the evaluation of CoT reasoning quality, similar to the 12.4% gap of Qwen-VL-Max. Conversely, SPHINX-MoE showcases favorable precision among open-source MLLMs, while preserving a relatively lower variance of two evaluation methods, i.e., 6.0% compared to InternLM-XComposer's 9.6%. This indicates its consistent step-by-step reasoning throughout the problem-solving process.

### 3.3 Error Analysis

To delve into the fine-grained predictions, we select the best-performing MLLM, GPT-4V [47], to understand its modes of success and failure. Our proposed CoT evaluation strategy has produced a detailed assessment of model output, including step-wise scores and explanation, reducing extensive manual effort in identifying and analyzing errors. We conduct our analysis on the two-step output from the CoT evaluation across the entire dataset, focusing on two key dimensions.

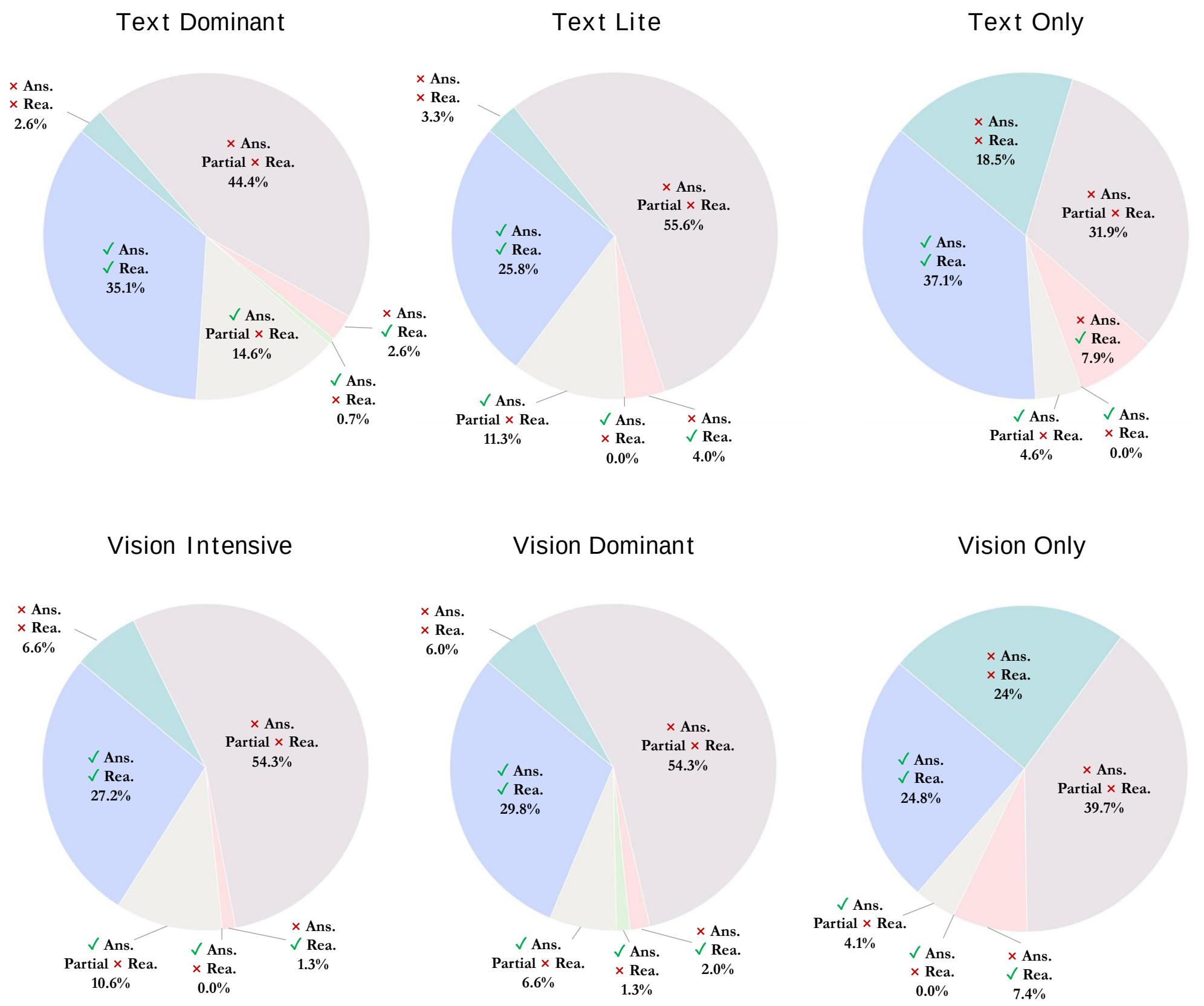

Figure 9: **Distribution of GPT-4V's [47] Errors in Reasoning and Answers.** For the six problem versions in MATHVERSE, we provide the statistics of errors made by GPT-4V based on their occurrence in answers ('Ans.') and reasoning processes ('Rea.').

**Errors in Reasoning or Answer?** In Figure 9, we showcase the statistics of different error distributions in six problem versions of MATHVERSE. We define the following six error categories: correct final answer with correct/partially correct/incorrect CoT reasoning and incorrect final answer with correct/partially correct/incorrect CoT reasoning. For all six versions, the incorrect final answers are mostly caused by the partially incorrect reasoning process. In addition, a number of problems with correct answers are accompanied by partially or entirely incorrect reasoning, e.g., 15.3% in Text Dominant, which cannot be detected by the traditional True or False evaluation. As we remove the content within textual questions and enrich the visual diagram, e.g., from Text Dominant and Lite to Vision Dominant and Only, we observe a progressive increase in the error rate of 'incorrect final answer with incorrect CoT reasoning', indicating that MLLMs are challenged to conduct high-quality intermediate reasoning by capturing more information from the visual input.

**What Types of Errors?** To further investigate the specific error types, we survey the problems with errors that occur either within the reasoning process or the final answer. As depicted in Figure 10, we divide the errors of GPT-4V into four distinct types: visual perception error, reasoning error, knowledge error, and calculation error. Consistent with our findings in the main paper, the primary source of errors in problem-solving attributes to the inaccurate interpretation of mathematical diagrams, which significantly impedes the performance of MLLMs. For the problem versions that demand advanced diagram interpretation, e.g., Vision Dominant and Only, we observe a notable increase in the rate of visual perception errors, demonstrating an urgent need for stronger visual encoders in MLLMs. Moreover, reasoning errors also account for a considerable percentage, indicating that the logical deduction skills of MLLMs still require improvement. As expected, knowledge errors do not significantly hinder the mathematical reasoning capabilities of MLLMs in MATHVERSE.

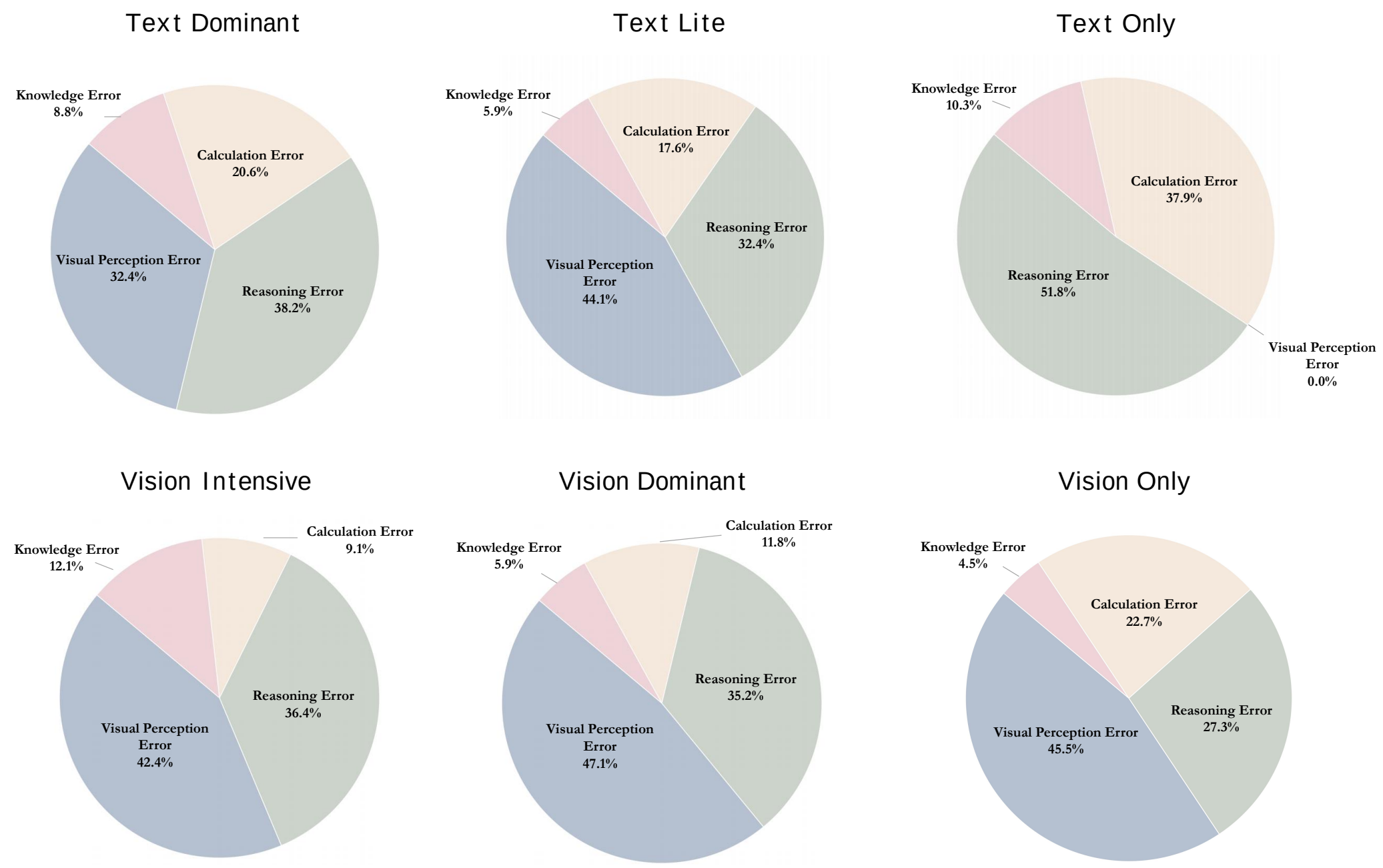

Figure 10: **Distribution of GPT-4V's [47] Errors within Different Types.** We present the statistics of four error types by GPT-4V in the six problem versions, i.e., Visual Perception Error, Reasoning Error, Calculation Error, and Knowledge Error.

# 4 Conclusion

In this paper, we propose a comprehensive and specialized benchmark, MathVerse, for the visual mathematical problem-solving capacity of MLLMs. We meticulously collect high-quality math problems with diagrams spanning three primary subjects and twelve subfields. Given the issues within current benchmarks, we transform each problem into six versions, investigating whether and how much MLLMs can interpret the visual math diagrams. We also propose a CoT evaluation strategy for finer-grained assessment of the intermediate reasoning process of MLLMs. By evaluating various closed-source and open-source models, MathVerse unveils that most existing MLLMs struggle to accurately understand mathematical diagrams, and even attain higher results without visual input. This indicates the potential of developing more advanced math-specific vision encoders for stronger multi-modal mathematical reasoning.

## Appendix Overview

- Section A: Related work.
- Section B: Additional experimental details.
- Section C: More dataset details.
- Section D: Comparison to current benchmarks.
- Section E: Limitation and future work.
- Section F: Qualitative examples.

# A Related Work

**Multi-modal Large Language Models (MLLMs),** building upon the prevalence of Large Language Models (LLMs) [55, 56, 45, 28, 4] and large vision models [49, 29, 68, 67, 69], have become increasingly prominent in the field. They extend LLMs to tackle a diverse range of tasks and domains, including the mainstream 2D images [35, 15, 1, 31] and other modalities, such as 3D point clouds [23, 60, 27], audio [24, 52], and video [65, 6]. Noteworthy examples like OpenAI's GPT-4V [47] and Google's Gemini [22] exhibit exceptional visual understanding and reasoning capabilities, setting new benchmarks in multi-modal performance. However, their closed-source nature poses a barrier to the broader application and development of MLLMs. Concurrently, another line of work is dedicated to exploring advanced MLLMs open-source to the community. Prior efforts like LLaMA-Adapter [66, 20], LLaVA [39, 38, 37], and MiniGPT-4 [72, 11] leverage a frozen CLIP [49] model for image encoding, and inject the visual cues into LLaMA [55] for multi-modal instruction tuning. The subsequent mPLUG-Owl [61, 62], Qwen-VL [3], InternLM-XComposer [16], and SPHINX [36, 21] further push the frontier of MLLMs in understanding and generalizing across visual contexts. Despite comprehensive benchmarks [17, 40, 33, 59] on general visual instruction-following scenarios, the specific potential of MLLMs for visual mathematical problem-solving remains under-explored. In this paper, we introduce the MATHVERSE benchmark to comprehensively evaluate the visual mathematical reasoning and diagram understanding skills of MLLMs, providing unique perspectives for future research directions.

**Mathematical Reasoning Benchmarks** have emerged as a significant area of focus, posing considerable challenges for large foundational models, e.g., LLMs and MLLMs. Initially, datasets in this realm are designed to address basic algebraic [26] and arithmetic [50] word problems, which are relatively limited in scope and volume. Subsequent efforts, including MATH [26], GSM8K [14], and MMLU [25], expand the range and quality of textual mathematical problems. These datasets feature a broader spectrum of difficulties, establishing a robust benchmark for the evaluation of general and math-specific LLMs [71, 64, 57, 19, 44]. Besides the text-only assessment, there is a growing demand for comparable, high-quality benchmarks for evaluating mathematical problem-solving in visual contexts, with the rapid progress of MLLMs. There are prior attempts, such as GeoQA [10], UniGeo [8], and Geometry3K [43], which focused exclusively on geometric problems. The recently proposed MathVista [41] broadens the scope to incorporate a variety of multi-modal tasks involving mathematical reasoning, and MMMU [63] covers college-level questions demanding intricate, domain-specific knowledge. However, our analysis identifies three main shortcomings within the current visual math benchmarks, as elaborated in Section 1 of the main paper. Therefore, we propose MATHVERSE specialized in the multi-modal mathematical evaluation of MLLMs, comprising twelve subjects, six problem versions, and 20K test samples. Our objective is to thoroughly investigate whether and how much MLLMs genuinely interpret visual diagrams for mathematical reasoning.

# B Additional Experimental Details

**Model Sources.** For different MLLMs, we select their latest models and best-performing configurations for evaluation to fully reveal their visual mathematical proficiency. Table 4 presents the release time and model sources of MLLMs used in MATHVERSE.

Table 4: **The Release Time and Model Source of MLLMs Used in MathVerse.**

| Model | Release Time | Source |
|---|---|---|
| ChatGPT [48] | 2022-11 | https://platform.openai.com/ |
| GPT-4 [46] | 2023-03 | https://platform.openai.com/ |
| Qwen-VL-Plus [3] | 2023-11 | https://help.aliyun.com/zh/dashscope/developer-reference/vl-plus-quick-start |
| Gemini-Pro [22] | 2023-12 | https://ai.google.dev/ |
| Qwen-VL-Max [3] | 2024-01 | https://help.aliyun.com/zh/dashscope/developer-reference/vl-plus-quick-start |
| GPT-4V [47] | 2023-09 | https://platform.openai.com/ |
| LLaMA-Adapter V2 [20] | 2023-04 | https://github.com/OpenGVLab/LLaMA-Adapter/tree/main/llama_adapter_v2_multimodal7b |
| LLaVA-1.5 [37] | 2023-10 | https://huggingface.co/liuhaotian/llava-v1.5-13b |
| MiniGPT-v2 [11] | 2023-10 | https://github.com/Vision-CAIR/MiniGPT-4 |
| mPLUG-Owl2 [62] | 2023-11 | https://huggingface.co/MAGAer13/mplug-owl2-llama2-7b |
| G-LLaVA [19] | 2023-12 | https://github.com/pipilurj/G-LLaVA/tree/main |
| ImageBind-LLM [24] | 2023-05 | https://github.com/OpenGVLab/LLaMA-Adapter/tree/main/imagebind_LLM |
| ShareGPT4V [12] | 2023-11 | https://huggingface.co/Lin-Chen/ShareGPT4V-13B |
| SPHINX-Plus [36] | 2023-11 | https://huggingface.co/Alpha-VLLM/LLaMA2-Accessory/tree/main/finetune/mm/SPHINX/SPHINX-v2-1k |
| LLaVA-NeXT [38] | 2024-01 | https://huggingface.co/liuhaotian/llava-v1.6-vicuna-13b |
| SPHINX-MoE [21] | 2024-01 | https://huggingface.co/Alpha-VLLM/LLaMA2-Accessory/tree/main/finetune/mm/SPHINX/SPHINX-MoE |
| InternLM-XComposer2 [16] | 2024-01 | https://huggingface.co/internlm/internlm-xcomposer2-vl-7b |

Table 5: **Input Prompt of MLLMs for Response Generation.** We adopt two different prompts for the free-form and multiple-choice questions. Note that these prompts are used for five problem versions except for the Vision-only version.

| Question | Prompt |
|---|---|
| Free-form Question | Please first conduct reasoning, and then answer the question and provide the final value, e.g., 1, 2.5, 300, at the end.<br>– **Question:** $\{question\}$ |
| Multiple-choice Question | Please first conduct reasoning, and then answer the question and provide the correct option letter, e.g., A, B, C, D, at the end.<br>– **Question:** $\{question\}$ |

Table 6: **Input Prompt for Vision-only Problems.** Especially for the Vision-only version without textual input, we add *"According to the question shown in the image"* at the beginning of the prompt, and remove the *"Question:"* at the end.

| Question | Prompt |
|---|---|
| Free-form Question | According to the question shown in the image, please first conduct reasoning, and then answer the question and provide the final value, e.g., 1, 2.5, 300, at the end. |
| Multiple-choice Question | According to the question shown in the image, please first conduct reasoning, and then answer the question and provide the correct option letter, e.g., A, B, C, D, at the end. |

**Prompt for Response Generation.** We adopt two types of prompts respectively for the free-form and multiple-choice questions, as shown in Table 5. We inspire the Chain-of-Thought (CoT) reasoning capabilities of MLLMs by using the phrase *"first conduct reasoning"*. Especially for the Vision-only problem version in Table 6, we add *"According to the question shown in the image"* at the beginning to remind MLLMs to read the questions rendered within diagrams, where the textual input for MLLMs only contains the prompt itself.

**Prompt for the CoT Evaluation.** Our proposed CoT evaluation contains two steps, i.e., key-step extraction and multi-step scoring, which prompt GPT-4 [46] and GPT-4V [47], respectively. The input configuration is listed in Table 7. We utilize the text-only GPT-4 in the first step to extract multiple key steps within the model's unstructured output, without feeding the question information. In the second step, we input the extracted key-step reasoning and all the available content related to the problem into GPT-4V, allowing for a holistic assessment, including diagram interpretation, logical reasoning, and numerical computation. In Figure 11, we showcase the manual annotation for critical information within functional diagrams, e.g., function expression and properties. This assists GPT-4V in evaluating the visual perception accuracy of MLLMs for function graphs.

**Human Performance Assessment.** We recruit ten qualified college students specifically for the evaluation of human performance on MATHVERSE. These individuals are kept separate from the data curation stage, eliminating the possibility of them encountering the solutions beforehand. We allocate to each student the questions from a specific problem version. This strategy is to prevent them from gaining additional information from another version to answer questions, e.g., leveraging the textual *Implicit Property* from the Text-lite version to solve Text-intensive problems. They are asked to directly provide the final answer without detailed reasoning. Therefore, we do not report the CoT evaluation results for human performance, alongside the 'Random Chance' baseline.

Table 7: **Configuration for the CoT Evaluation Strategy.** We conduct two evaluation phases, respectively by prompting the text-only GPT-4 [46] and GPT-4V [47]. The symbol 'XXX' denotes the given one-shot sample, abbreviated for brevity. The 'Annotation' (represented in grey) in the second phase is only required for function problems.

| Phase | Input | Prompt |
|---|---|---|
| **Key-step Extraction** (GPT-4) | Model Output | I will give you a detailed solving procedure or a single answer for a math problem.<br>If it is a procedure, you need to extract the key solution steps and list them accordingly in markdown syntax. If it is just a single answer, output the answer directly.<br>Here are examples:<br>– Model output: XXX<br>– Extracted: 1. XXX 2. XXX 3. XXX<br>– Model output: 2.2<br>– Extracted: The single answer is 2.2<br>Here is what you need to extract:<br>– **Model output:** $\{model\ output\}$<br>– **Extracted:** |
| **Multi-step Scoring** (GPT-4V) | Extracted Steps<br>Question<br>Diagram<br>Answer<br>Annotation | I will first give you a visual math problem, including the question, diagram, ground-truth answer, and detailed annotation of the diagram, and then give you a model output containing multiple key solution steps.<br>Please think step by step and output the Average score, along with the Final answer score in the end, as described below:<br>– Average score: Evaluate, based on the given question, answer, diagram, and diagram annotation, whether each solution step is correct in logical reasoning, visual perception, and numerical computation, with an incorrect score of 0 and a correct score of 1. Then, calculate the average score of multiple steps.<br>– Final answer score: Match the model's final answer with the ground truth answer, scoring 1 if it matches and 0 if it doesn't.<br>– If the model output only includes a single step or answer, the Average score and Final answer score are the same.<br>– **Question:** $\{question\}$<br>– **Ground-truth answer:** $\{answer\}$<br>– **Diagram annotation:** $\{annotation\}$<br>– **Model output:** $\{extracted\ steps\}$<br>– **Average score:**<br>– **Final answer score:** |

**Annotation:**

A piecewise function $f$:
when $x <= 3,\ f(x) = -x + 1$,
when $x > 3,\ f(x) = x - 5$.

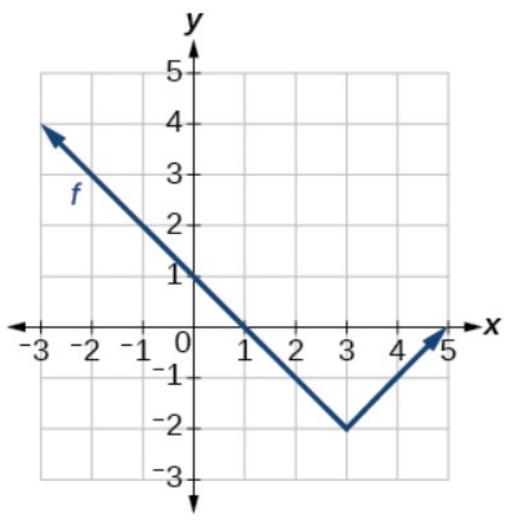

**Annotation:**

A U-shape curve,
passes the origin (0, 0),
opens upward.

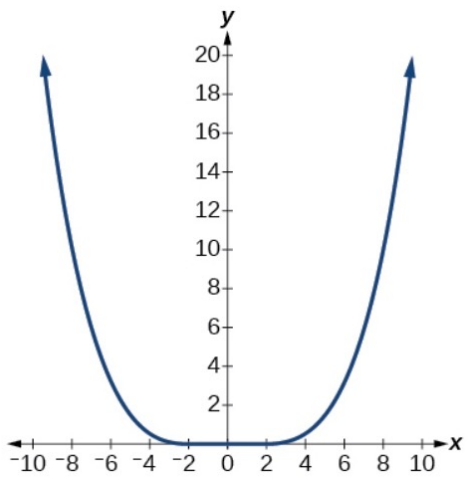

**Annotation:**

A quadratic function $y = (x - 2)^2 - 4$,
goes through (0, 0), (4, 0),
opens upward with a vertex (2, 4).

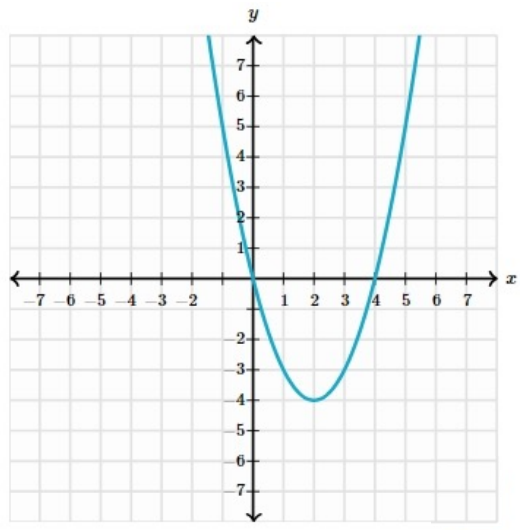

**Annotation:**

Two quadratic functions going through (1, 0) and (7, 0),
$f(x) = -(1/3)(x - 1)(x - 7)$ in blue with a vertex is (4, 3),
$g(x) = -(2/3)(x - 1)(x - 7)$ in dashed red with a vertex is (4, 6).

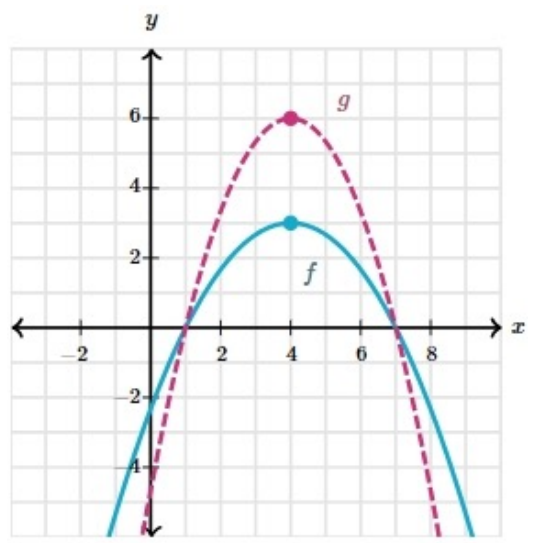

**Annotation:**

A linear function $y = 0.5x + 2$,
passes (-4, 0), (0, 2) and (2, 3),
the shaded region is above the line.

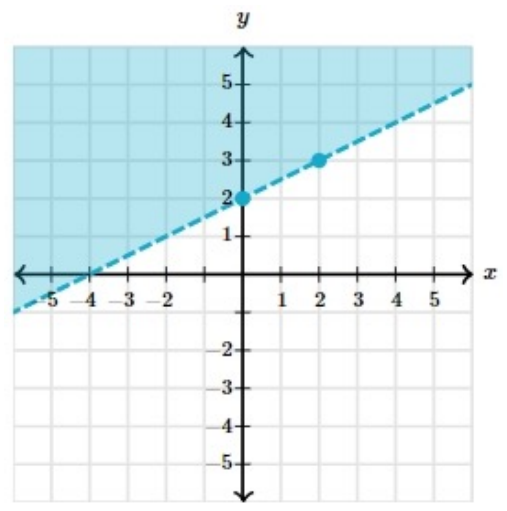

Figure 11: **Manual Annotations for Function Problems in MathVerse.** We provide detailed annotations, e.g., function expression and properties, for the diagrams of 534 function problems, which benefits the accuracy of GPT-4V [47] for CoT evaluation.

# C More Dataset Details

## C.1 Data Curation

This paper engages twelve expert annotators for data curation, consisting of senior undergraduate and graduate students from across the globe with a strong background in science. In collaboration with the authors, they are required to mainly complete five tasks concerning data collection, categorization, quality review, problem version transformation, and function diagram annotation.

**Data Collection.** We comprehensively collect visual math problems from existing datasets [43, 9, 51] and public question repositories[1,2,3]. We specifically select high-quality plane geometric problems from current benchmarks, which showcase various question types, moderate question length, diverse diagram styles, and appropriate solving difficulty. For the manually collected problems of three subjects (plane geometry, solid geometry, and functions), we apply the Mathpix tool[4] to accurately extract the question texts, diagrams, explanations, and answers from the website. We strictly comply with copyright and licensing rules, ensuring that we refrain from using data from sites that forbid copying and redistribution. After the initial collection, we obtain around 3.5K visual math problems, with 1.5K from existing datasets and 2K newly collected.

[1] https://homework.study.com
[2] https://www.ixl.com/math
[3] https://mathspace.co/us
[4] https://mathpix.com

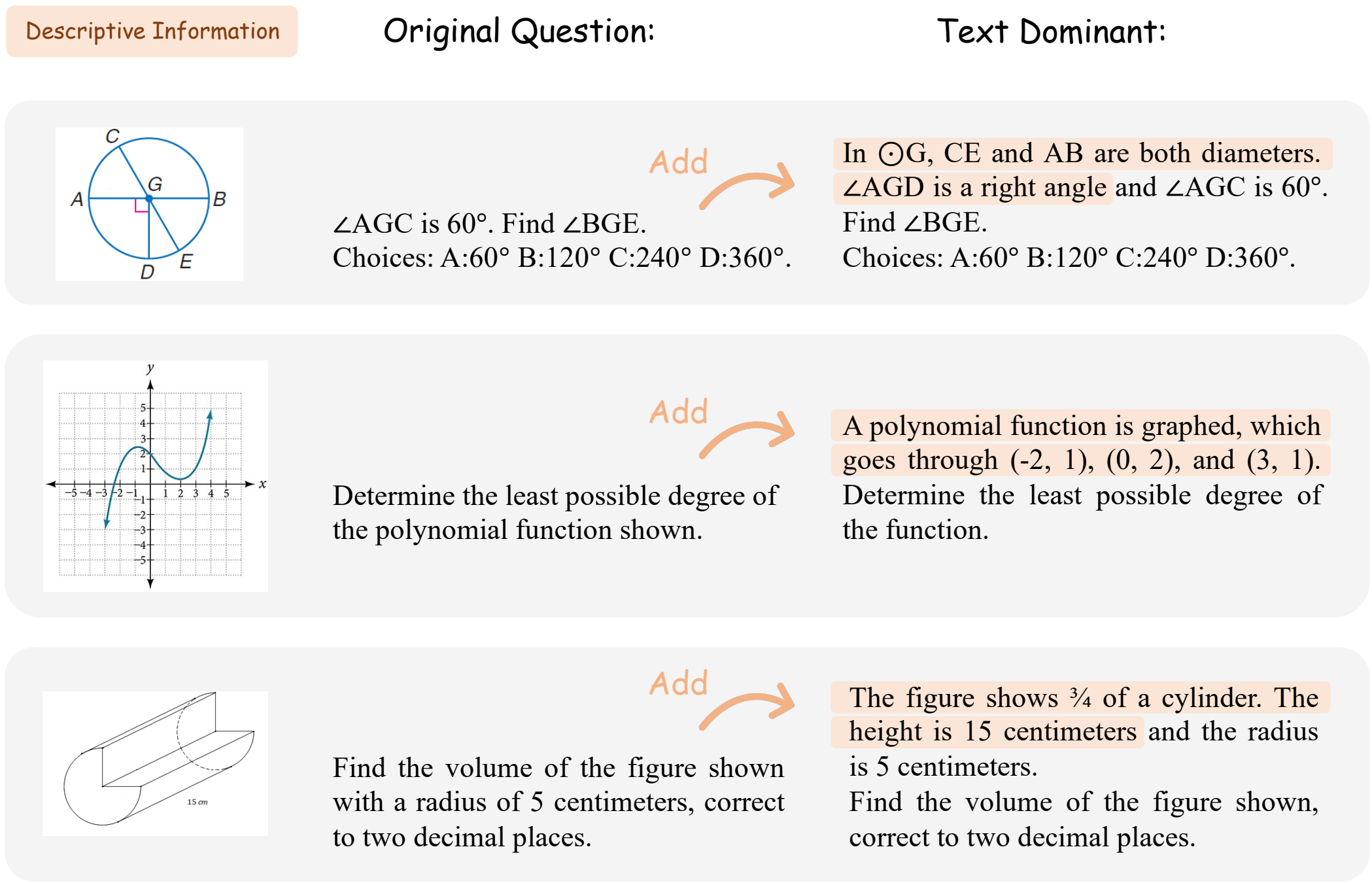

Figure 12: **Manual Annotations for *Descriptive Information* in MathVerse.** For some collected problems, we are required to supplement additional *Descriptive Information* (highlighted in red) to distinguish the Text-dominant version.

**Data Categorization and Review.** We first ask the human annotators to categorize the problems into three primary subjects, i.e., plane geometry, solid geometry, and functions. Within each subject, according to the definitions in Section C.2, the math problems are further divided into twelve fine-grained categories. At the same time, we meticulously review the collected dataset. We manually rectify the problems with incorrect answers and discard the problems with multiple diagrams, visual solutions, and too much similar content to others. Finally, 2,612 high-quality math problems with paired diagrams are preserved for MathVerse, spanning diverse subjects and subfields.

**Transformation of Problem Versions.** Given the three types of textual information within questions, human annotators rigorously transform each problem into six different versions as discussed in Section 2.2 of the main paper. We utilize Microsoft PowerPoint to annotate the diagrams in the Vision-dominant version, and employ Matplotlib to render the questions onto the diagrams in the Vision-only version. As illustrated in Figure 12, for problems with minimal *Descriptive Information*, we manually enhance the question text with additional contextual description about the diagram to differentiate the Text-dominant version. In the case of questions in Figure 13, where the *Essential Condition* has been fully depicted in the diagrams, we remove some of this content from the diagram and incorporate it into the text to mark the Vision-dominant version.

## C.2 Subject and Subfield Definition

The visual math problems within MathVerse encompass three primary subjects, plane geometry, solid geometry, and functions, alongside twelve finer-grained subfields, which comprehensively evaluate the diagram understanding and mathematical reasoning capabilities of MLLMs.

**Plane Geometry** is a fundamental area that explores the properties and relations of points, lines, and surfaces in a two-dimensional plane. This subject delves into concepts such as angles, triangles, circles, and polygons, offering a rich context for assessing the spatial comprehension and logical deduction skills of MLLMs. We divide it into five subfields, as exemplified in Figure 14:

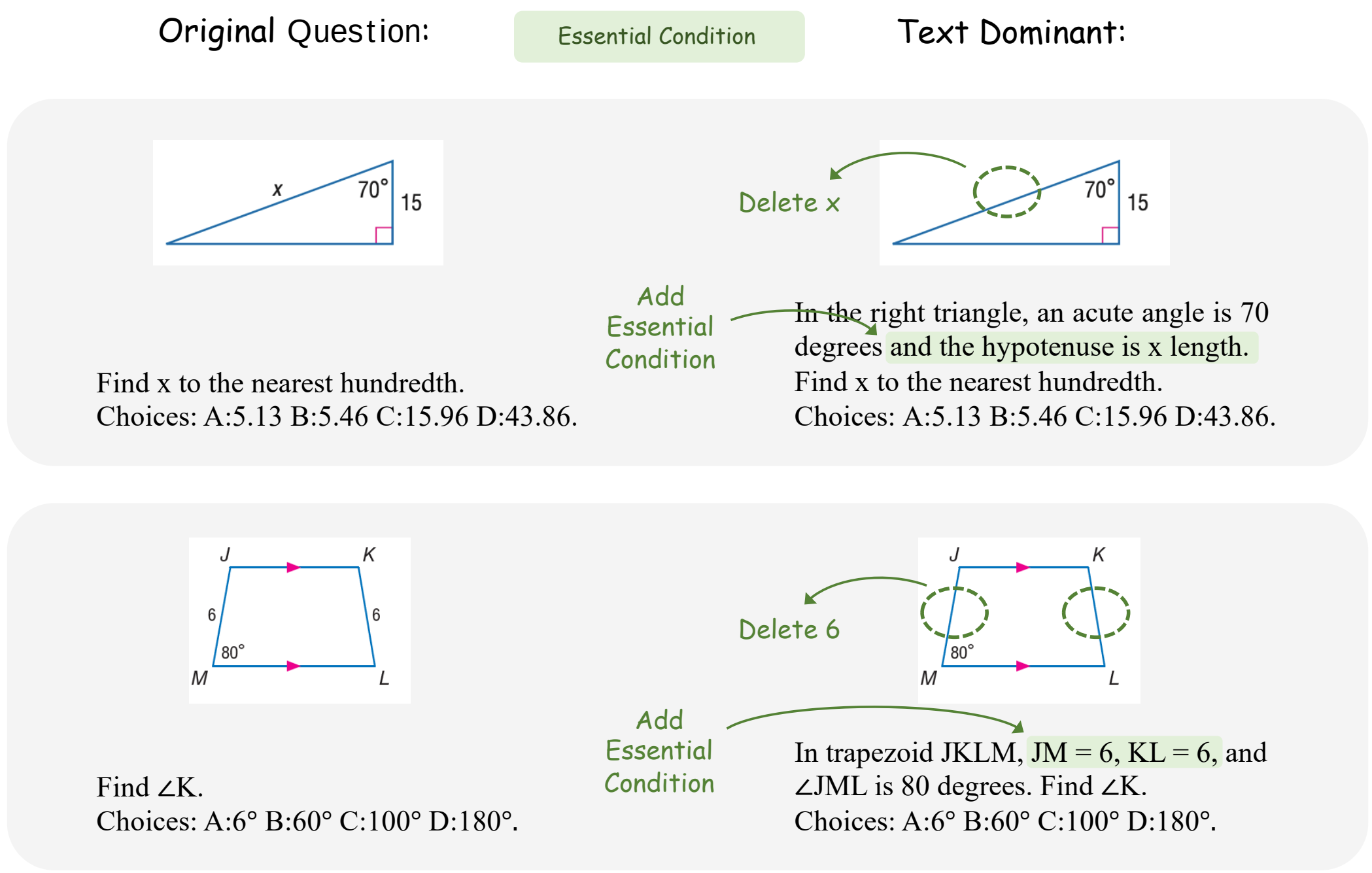

Figure 13: **Manual Modification for Textual *Essential Condition* in MATHVERSE.** For the original problems shown, we transfer some of the *Essential Condition* from diagrams to question texts (highlighted in green) to mark the Vision-dominant version.

- **Length** focuses on the measurement and comparison of distances between points. This subfield includes understanding the properties of lines, segments, and their use in determining the perimeters of geometric shapes, which is foundational for MLLMs to solve plane geometry problems.
- **Area** examines the size of two-dimensional surfaces. It encompasses calculating the areas of various shapes, such as triangles, rectangles, circles, and more complex polygons, by applying specific formulas and principles, which is crucial for comprehending the concept of space within geometry.
- **Angle** involves the study of angles and their properties, including different types of angles (acute, right, and obtuse), angle measurement, and the relationships between angles, particularly in polygons. This subfield demands the advanced spatial perception capacity of MLLMs.
- **Analytic Geometry**, also known as coordinate geometry, merges algebra and geometry to solve geometric problems using coordinate systems, exploring the calculation and reasoning of equations for geometric shapes. MLLMs are evaluated on their coordinate identification and algebraic capabilities.
- **Applied Geometry** relate to the application of geometric principles to solve real-world and theoretical problems. It challenges MLLMs to first understand the background information within questions, and apply their knowledge of lengths, areas, angles, and analytic geometry for problem-solving.

**Solid Geometry** focuses on the study of three-dimensional objects that have depth, length, and width, thereby offering a more complex and enriched exploration of spatial structures. This subject investigates a variety of shapes such as cubes, cylinders, spheres, and pyramids, and assesses MLLMs to tackle questions concerning the volume, surface area, and geometric properties of these solids. This subject contains three subfields, as exemplified in Figure 15:

- **Length**, extending from the 2D counterpart, focuses on measuring the edges and curves that define three-dimensional objects. It involves determining the linear distance between points in space, the perimeters of bases of solids, and the height or depth of objects. This measurement is a foundational element for MLLMs in analyzing geometric solids.

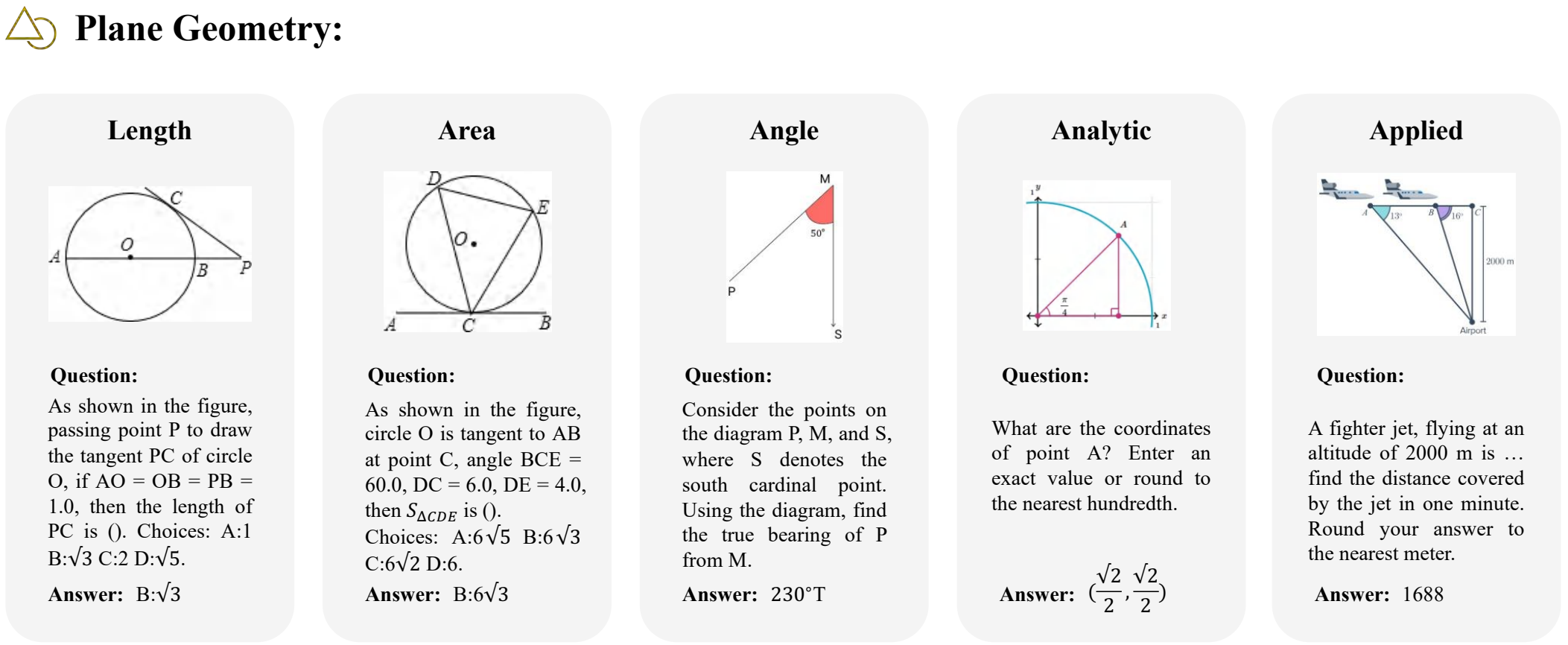

Figure 14: **Examples of Five Subfields in Plane Geometry**, spanning Length, Area, Angle, Analytic, and Applied Geometry problems. We showcase the Text-lite version.

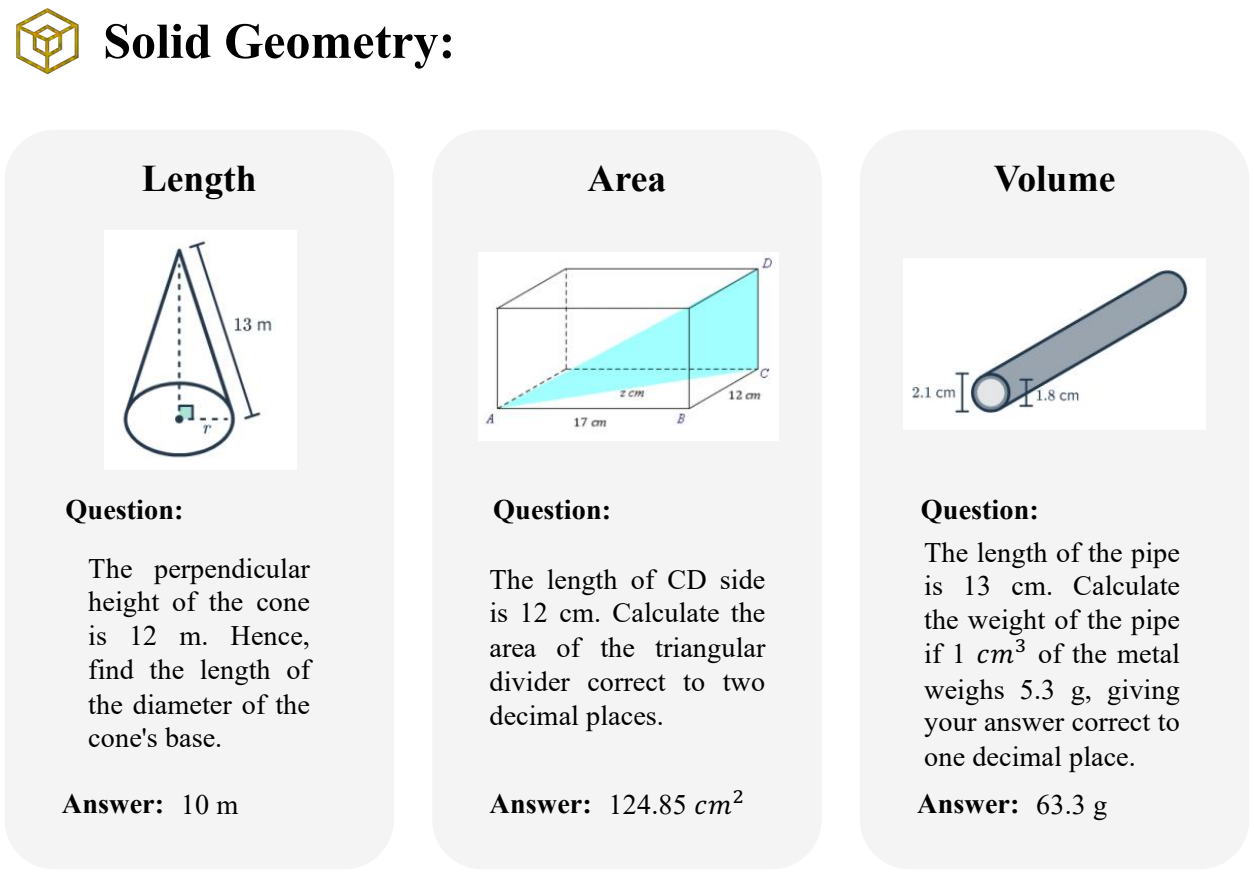

Figure 15: **Examples of Three Subfields in Solid Geometry**, spanning Length, Area, and Volume problems. We showcase the Text-lite version.

- **Area** encompasses the calculation of the total area covered by the outer surfaces of solids. This normally requires MLLMs to break down complex shapes into several simpler components for area calculation in plane geometry, assessing their spatial and logical reasoning performance.
- **Volume** pertains to measuring the space enclosed within three-dimensional objects. This demands MLLMs to precisely identify the geometric solids and apply accurate formulas to calculate the volume, which evaluates their mathematical knowledge application and calculation skills.

**Functions** involve analyzing mathematical functions to understand the relationship between variables. These challenges range from simple tasks, like calculating a function value for a given input, to more complex scenarios, such as exploring the behavior and representation of various function types. We evaluate MLLMs by four types of function problems, exemplified in Figure 16:

- **Function Coordinate** focuses on interpreting and extracting coordinate-level information from graphical representations of functions. It includes tasks such as identifying specific coordinate values of points on the graph and observing intersection points between functions and axes, which test the MLLM's basic proficiency in functional visual perception.

**Functions:**

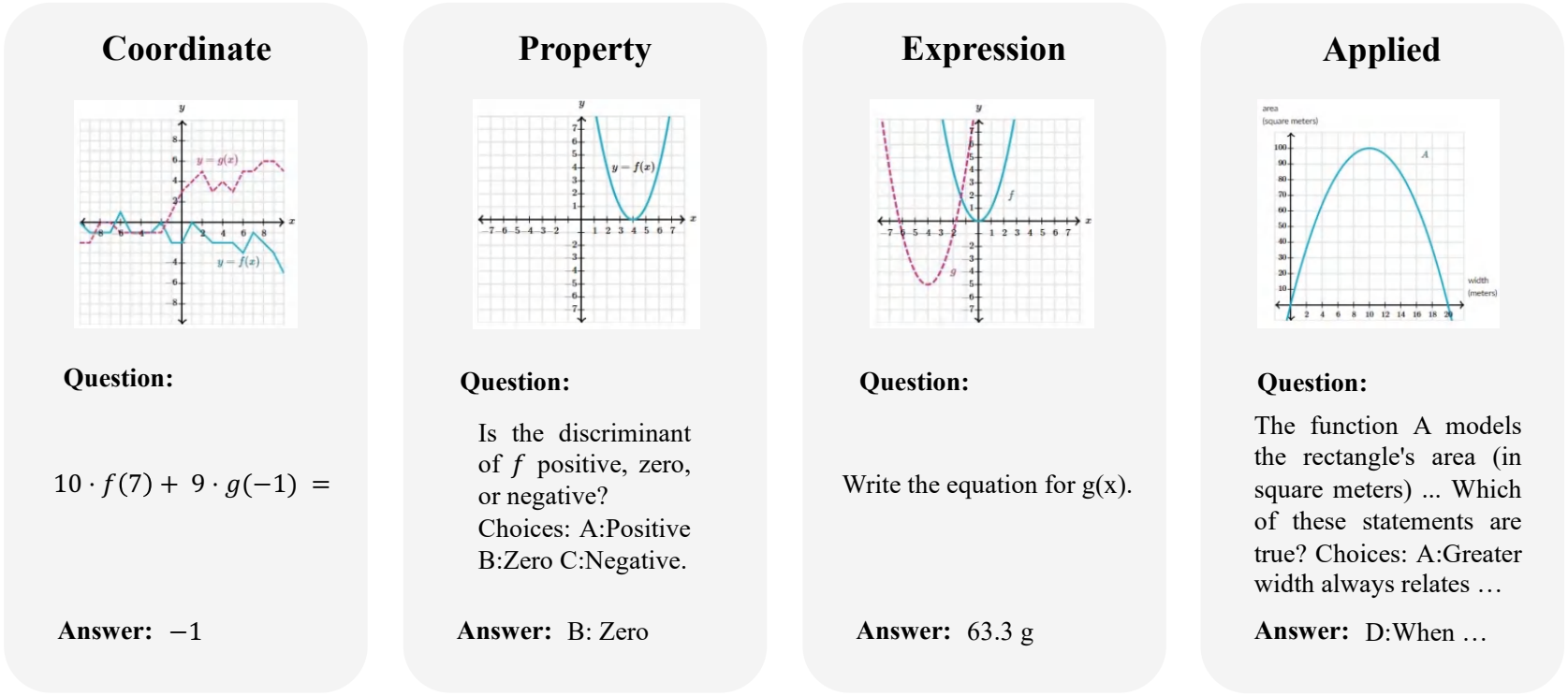

Figure 16: **Examples of Four Subfields in Functions**, spanning Function Coordinate, Property, Expression, and Applied problems. We showcase the Text-lite version.

- **Function Property** emphasizes the model's capacity to discern and deduce the inherent properties of functions from their graphs, such as symmetry, asymptotes, extrema (maximum and minimum points), and intervals of increase or decrease. These problems can reveal the understanding of MLLMs for the deeper characteristics of functions.

- **Function Expression** refers to the direct analysis using the algebraic expressions of functions, widely including linear, quadratic, polynomial, exponential, logarithmic, and piecewise functions. It challenges MLLMs to extract specific function expressions and apply transformations, bridging the gap between abstract mathematical reasoning and visual interpretation.

- **Applied Function**, similar to the applied geometry problems, requires MLLMs to leverage their functional knowledge and theorems in real-world scenarios, e.g., modeling economic data, predicting physical phenomena, and calculating probabilities. This assesses the MLLM's capabilities to understand functions in both theoretical and practical contexts.

### C.3 Detailed Statistics of MathVerse

**More Data Statistics.** In Table 8, we provide a more detailed data statistics of MathVerse. Therein, the 534 newly annotated questions refer to all the function problems, for which we meticulously annotate critical functional information, as depicted in Figure 11. The number of newly annotated diagrams represents the 5,224 math problems in the Vision-dominant and Vision-only versions. For these problems, we respectively integrate the *Essential Condition* and all textual content with the diagrams. We also list the numbers of multiple-choice answers, where A, B, C, and D are almost uniformly distributed.

**Problem Length Variance.** In Table 9, we highlight the variance in question and answer lengths across the five problem versions in MathVerse, excluding the Vision-only category due to its absence of text. For both word and character levels, as we remove the pre-defined textual elements (*Descriptive Information, Implicit Property*, and *Essential Condition*), the maximum and average lengths of questions decrease accordingly, while the answer lengths remain the same. In Figure 17, we visualize the word-level variation of question length for the three problem versions: Text Dominant (blue), Text Lite (green), and Vision Dominant (red). By progressively omitting *Descriptive Information* and *Essential Condition* from the Text-dominant version, we observe a clear downward trajectory for the question length distribution and their average values.

Table 8: **Statistics of MATHVERSE.**

| **Statistic** | **Number** |
|---|---|
| Total questions | 2,612 |
| - Subjects/subfields | 3/12 |
| - Multiple-choice questions | 1,631 (62.4%) |
| - Free-form questions | 981 (37.6%) |
| - **Newly collected questions** | **1,236 (47.3%)** |
| - Existing-dataset questions | 1,376 (52.7%) |
| - **Questions with explanations** | **1,236 (47.3%)** |
| - **Newly annotated questions** | **534 (20.4%)** |
| Multiple-choice question | |
| - Proportion of answer A | 585 (22.4%) |
| - Proportion of answer B | 828 (31.7%) |
| - Proportion of answer C | 703 (26.9%) |
| - Proportion of answer D | 444 (17.0%) |
| - Proportion of answer E&F | 52 (2.0%) |
| **Total test samples** | **15,672** |
| - **Newly annotated samples** | **10,448 (66.7%)** |
| - **Newly annotated diagrams** | **5,224 (33.3%)** |
| - Samples of each version | 2,612 (16.7%) |
| Number of unique images | 2,420 (92.6%) |
| Number of unique questions | 2,573 (98.5%) |
| Number of unique answers | 847 (32.4%) |

Table 9: **Length of Different Problem Versions in MATHVERSE.**

| **Problem Version** | **Word** | **Character** |
|---|---|---|
| Text Dominant & Text Only | | |
| - Maximum question length | 203 | 1,311 |
| - Maximum answer length | 17 | 102 |
| - Average question length | 35.7 | 204.8 |
| - Average answer length | 1.4 | 6.3 |
| Text Lite | | |
| - Maximum question length | 179 | 1,173 |
| - Maximum answer length | 17 | 102 |
| - Average question length | 22 | 133.8 |
| - Average answer length | 1.4 | 6.3 |
| Vision Intensive | | |
| - Maximum question length | 171 | 1,126 |
| - Maximum answer length | 17 | 102 |
| - Average question length | 18.8 | 116.8 |
| - Average answer length | 1.4 | 6.3 |
| Vision Dominant | | |
| - Maximum question length | 176 | 1,132 |
| - Maximum answer length | 17 | 102 |
| - Average question length | 17.6 | 123.5 |
| - Average answer length | 1.4 | 6.3 |

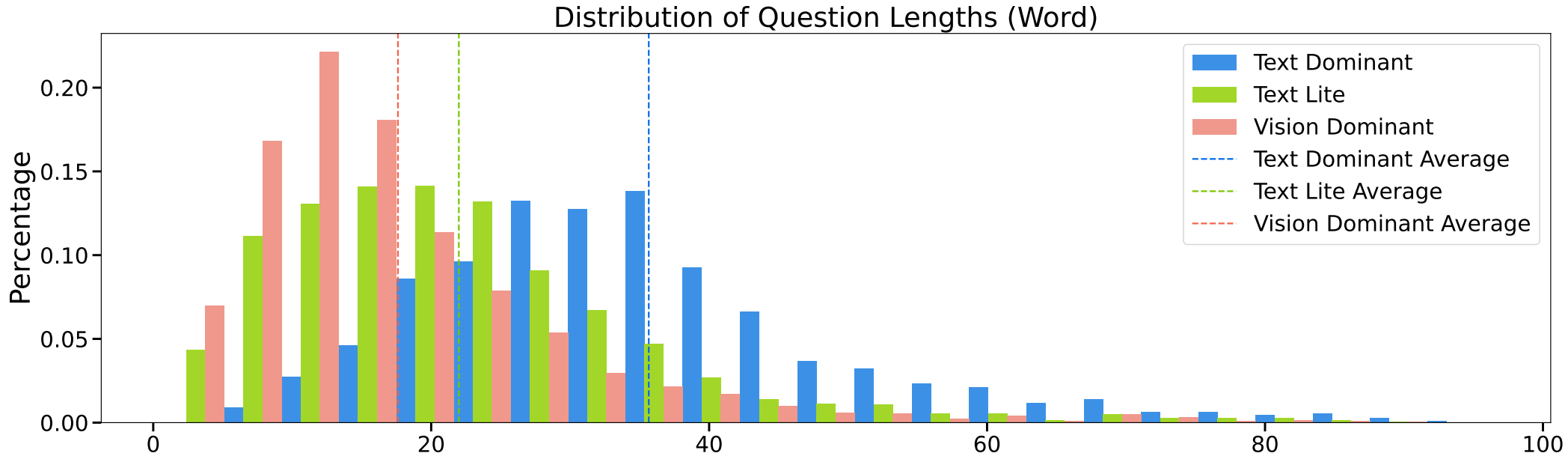

Figure 17: **Distribution of Question Length for Three Problem Versions.** We exclude the *Descriptive Information* and *Essential Condition* from the Text-dominant problems, respectively creating the Text-lite and Vision-dominant versions.

## D Comparison to Current Benchmarks

In this section, we offer a detailed comparison between MATHVERSE and existing multi-modal mathematical benchmarks, i.e., geometry-specific benchmarks [9, 5, 7, 42, 51], MathVista [41], and MMMU [63], from the following four aspects:

**The Investigation of Diagram Interpretation Capacity.** As discussed in Figure 1 of the main paper, the math problems in most existing datasets contain excessive redundant information in textual content, which is repetitive to the visual elements in diagrams. This issue enables MLLMs to potentially bypass the process of visual understanding, and thereby cannot determine whether and how much MLLMs truly interpret the math diagram. In contrast, our MATHVERSE includes six problem versions with different information content across text and vision. By comparing the performance variance between different problem versions, we can thoroughly investigate the mathematical diagram interpretation capabilities of MLLMs for the first time.

**Evaluation Approach.** Previous benchmarks adopt a simple True or False metric to score the response from MLLMs, which lacks fine-grained information and intermediate reasoning assessment, as analyzed in Figure 2 of the main paper. In contrast, MATHVERSE adopts a unique CoT evaluation

## Table QA:

| Items sold last week | |
|---|---|
| Item | Frequency |
| TV | 16 |
| tablet computer | 7 |
| speaker | 32 |
| cell phone | 23 |
| video game console | 19 |

| Making leaf rubbings | |
|---|---|
| Leaf rubbings made | Frequency |
| 0 | 14 |
| 1 | 5 |
| 2 | 15 |
| 3 | 11 |
| 4 | 3 |
| 5 | 10 |
| 6 | 17 |

| Rounds in the spelling bee | |
|---|---|
| Year | Rounds |
| 2008 | 17 |
| 2009 | 10 |
| 2010 | 9 |
| 2011 | 15 |
| 2012 | 7 |

Athletes per country

| Stem | Leaf |
|---|---|
| 1 | 1 3 3 |
| 2 | 0 1 2 5 7 9 9 |
| 3 | 4 5 7 8 |
| 4 | 1 |
| 5 | 2 4 5 5 6 6 8 |
| 6 | 1 6 |
| peanut butter cookie dough | $3 per lb |
| double chocolate cookie dough | $3 per lb |
| chocolate chip cookie dough | $5 per lb |
| oatmeal raisin cookie dough | $7 per lb |
| gingersnap cookie dough | $3 per lb |
| snickerdoodle cookie dough | $8 per lb |

## Textbook and Science QA:

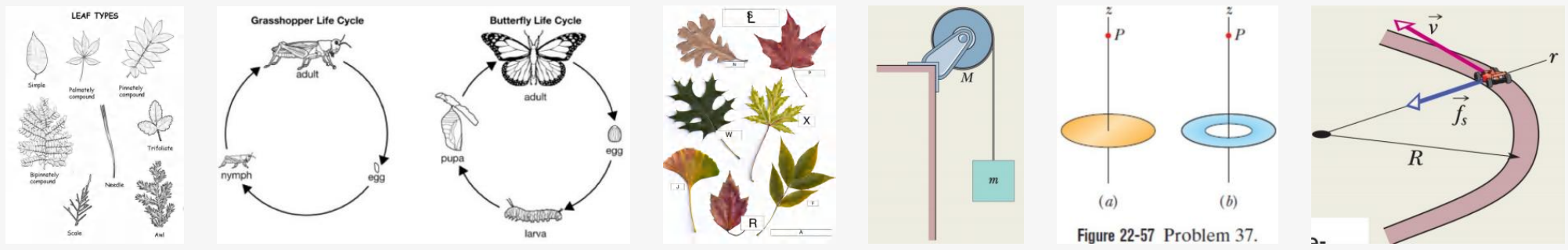

## Plot and Chart QA:

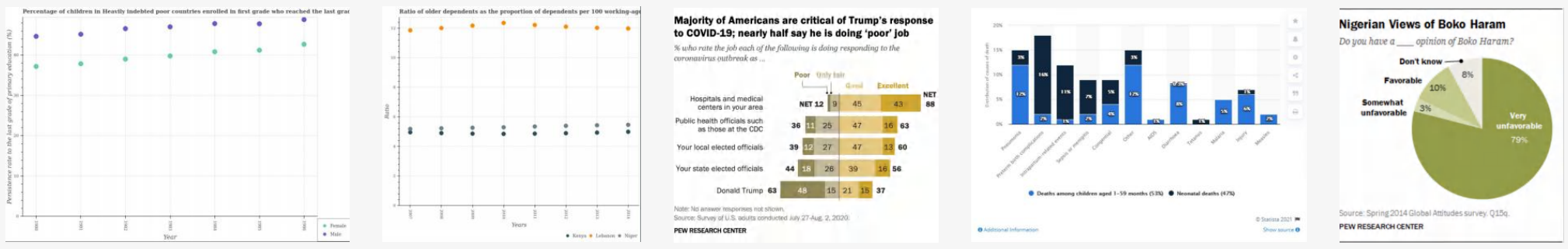

## IQ Test and Synthetic QA:

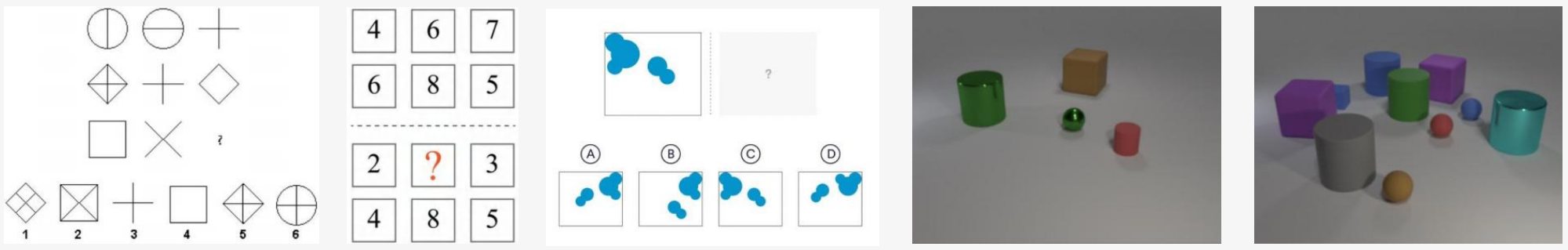

## General and Icon QA:

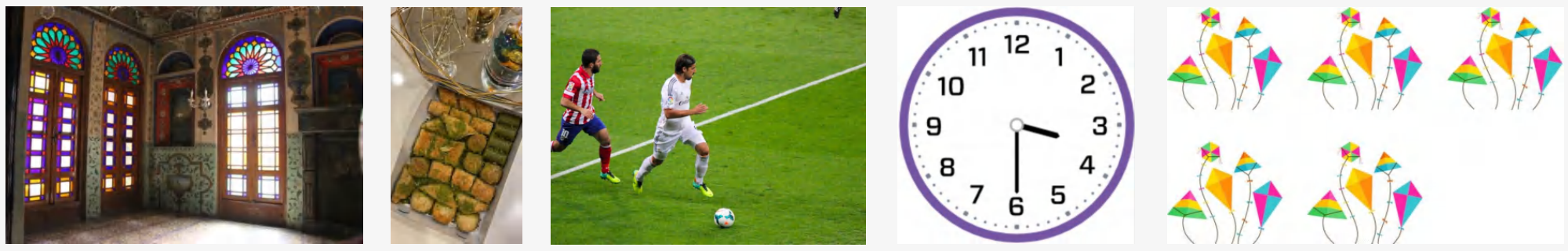

Figure 18: **Diagram Examples of Math-related Tasks in MathVista [41].** These tasks are not strongly correlated to the mathematical reasoning skills of MLLMs, probably skewing the assessment emphasis towards visual math problems.

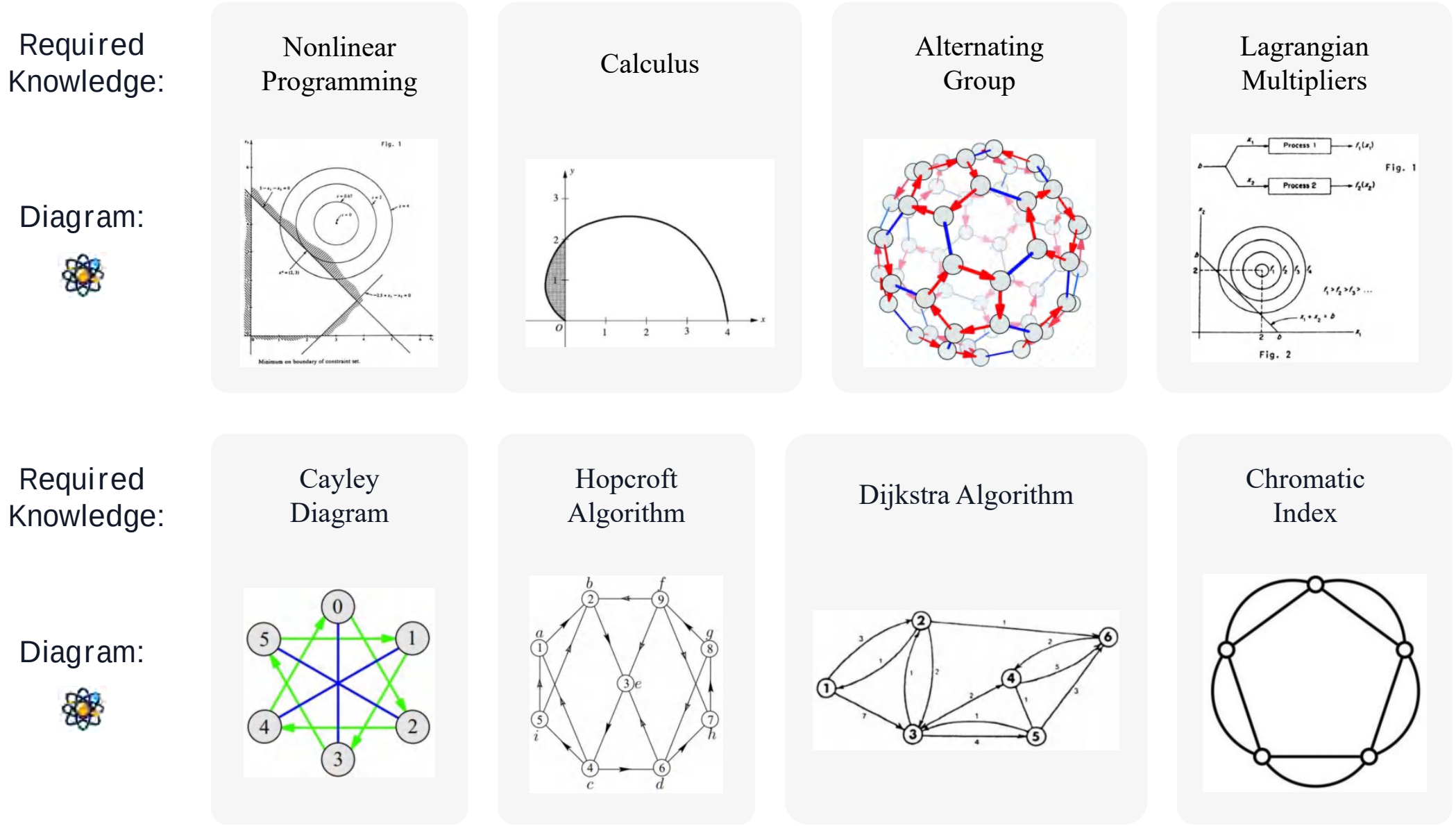

Figure 19: **Diagram Examples with Required Knowledge in MMMU [63].** These math problems demand MLLMs to comprehend college-level domain knowledge, potentially hindering them from fully exerting mathematical reasoning skills.

strategy by examining each crucial solution step within the model output. This approach not only unveils the CoT reasoning quality of MLLMs, but also provides detailed error analysis, serving as valuable guidance for future enhancement.

**The Depth and Width in Math Problems.** The geometry-specific benchmarks evaluate only a limited dimension of mathematical skills in MLLMs. MathVista instead incorporates a variety of math-related question-answering tasks, e.g., textbook figures, tables, plots, charts, puzzles, and synthetic scenes, as exemplified in Figure 18. However, the integration of these peripheral tasks (covering more than 70%) might divert the focus from the specialized mathematical evaluation of MLLMs. In addition, MMMU focuses on college-level complexity, requiring advanced domain-specific knowledge, as depicted in Figure 19. Given this, the lack of profound mathematical theorems would restrict the performance of MLLMs, biasing the evaluation of logical reasoning and visual perception proficiency. Therefore, our MATHVERSE concentrates on specialized visual math problems (plane geometry, solid geometry, and functions) with a moderate difficulty (high-school level), aiming to fully exert the capabilities of MLLMs.

**Total Volume of Test Samples.** We summarize the size of test instances for different datasets in Table 10. As demonstrated, our MATHVERSE offers a considerably larger number of samples than others, nearly three times to MathVista and twenty times to GeoQA+, including meticulously annotated six versions of visual math problems. This contributes to a comprehensive and robust evaluation of visual mathematical reasoning capabilities.

Table 10: **Number of Test Samples in Different Benchmarks.**

| **Benchmark** | GEOS | Geo3K | GeoQA+ | MathVista | MMMU-Math | **MATHVERSE** |
|---|---|---|---|---|---|---|
| **Test Samples** | 119 | 601 | 755 | 6,141 | 540 | **15,672** |

# E Limitation and Future Work

While our MathVerse takes a step forward in the field of visual mathematical evaluation for MLLMs, it is important to recognize several limitations as follows.

We have categorized the math problems in MathVerse by various criteria, including subjects, subfields, and versions featuring differing degrees of multi-modal content. These categorization approaches evaluate the capabilities of MLLMs from multiple dimensions. Nevertheless, it is also meaningful to further divide the problems based on their difficulty levels, akin to MATH [26], a text-only benchmark defining five levels of difficulty. This additional layer of differentiation can provide deeper insights into the problem-solving abilities of MLLMs across a spectrum of challenges, which we leave as future work.

The curated dataset in MathVerse focuses on math problems in the high school level with moderate difficulty, which aims to fully demonstrate the mathematical reasoning skills within current MLLMs. However, with the advancement of architecture and training methodologies, future MLLMs have the potential to grasp more complex knowledge and theorems across a variety of domains. Therefore, there is significant value in further augmenting MathVerse with problems spanning broader complexity and disciplines, including those at the college level and within scientific fields. By transforming the expanded problems into different versions, we can facilitate a more comprehensive and robust evaluation of MLLMs for their diagram interpretation and reasoning capabilities.

Moreover, the problems in MathVerse and other current mathematical benchmarks are mainly in English. Given that some multilingual MLLMs [24, 3] have been developed, existing evaluation cannot reveal their full capabilities when confined to a single language. The incorporation of multilingual visual math problems would not only extend the dataset's global applicability, but also enhance the assessment of MLLMs for linguistic diversity and understanding.

# F Qualitative Examples

To ease the understanding, we offer a variety of qualitative examples in MathVerse. In Section F.1, we showcase the meticulously transformed six versions of visual math problems. In Section F.2, we compare the response of different MLLMs on Text-lite problems, including GPT-4V [47], LLaVA-NeXT [38], and SPHINX-MoE [21]. Specifically, we present the key-step extraction output by the CoT evaluation, and mark the multi-step scoring results aside. In Section F.3, we provide the response comparison of GPT-4V for three problem versions in MathVerse, i.e., Text Dominant, Text Lite, and Vision Dominant.

## F.1 Comparison of Six Problem Versions

Please refer to Figures 13∼15.

## F.2 Response of Different MLLMs

Please refer to Figures 16∼21.

## F.3 Response of Different Problem Versions

Please refer to Figures 22∼27.

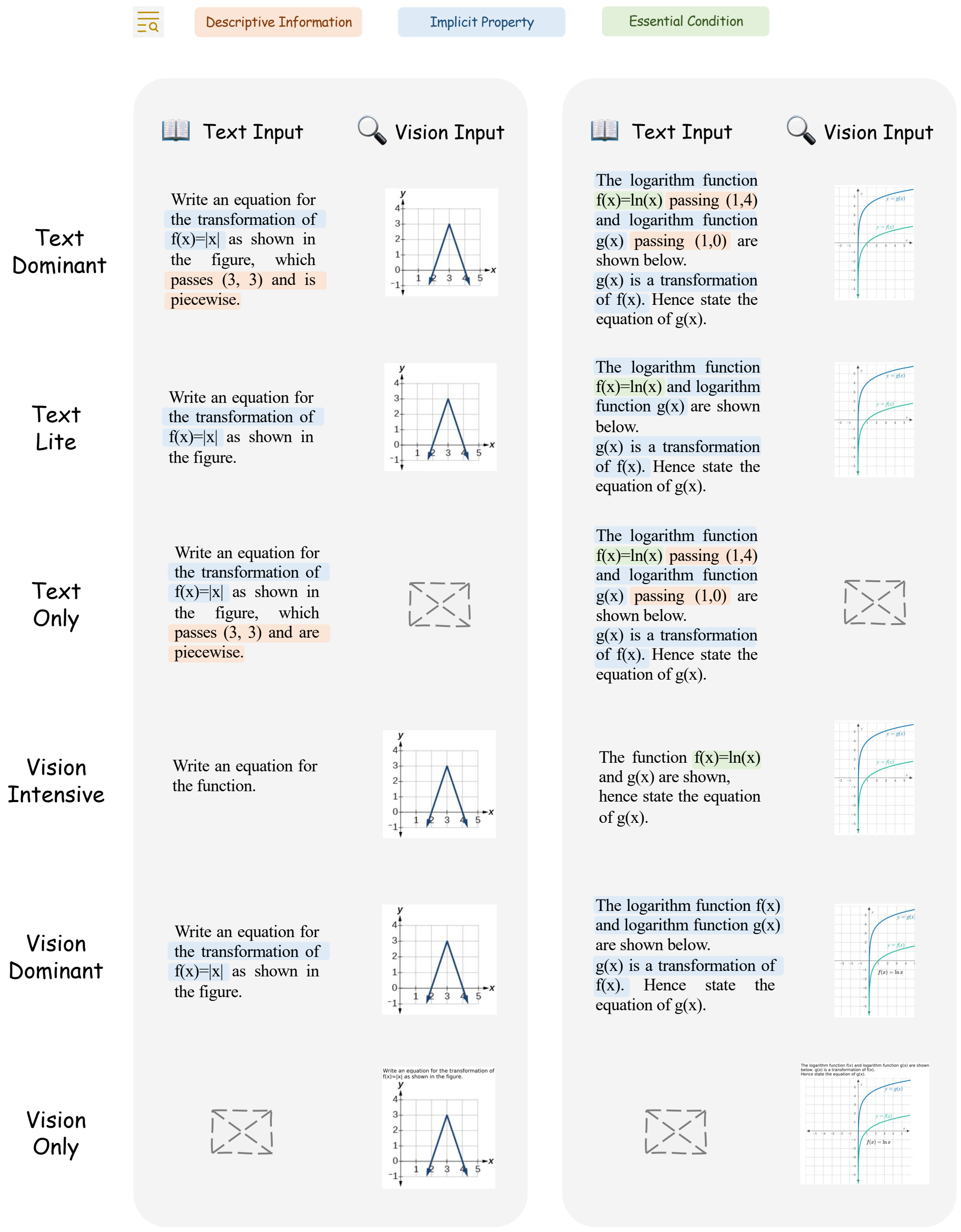

Figure 20: **Comparison of Six Problem Versions in MathVerse.**

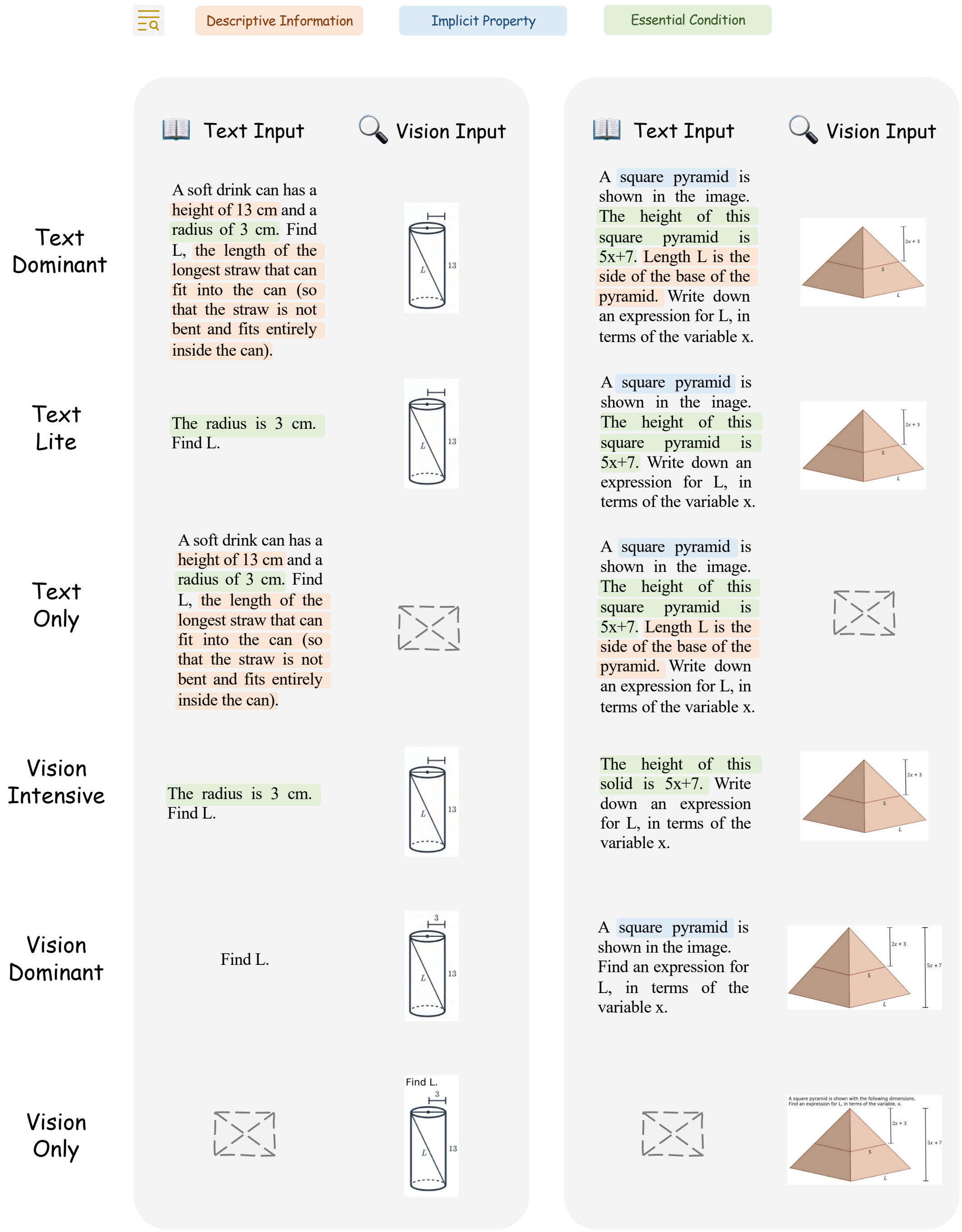

Figure 21: **Comparison of Six Problem Versions in MATHVERSE.**

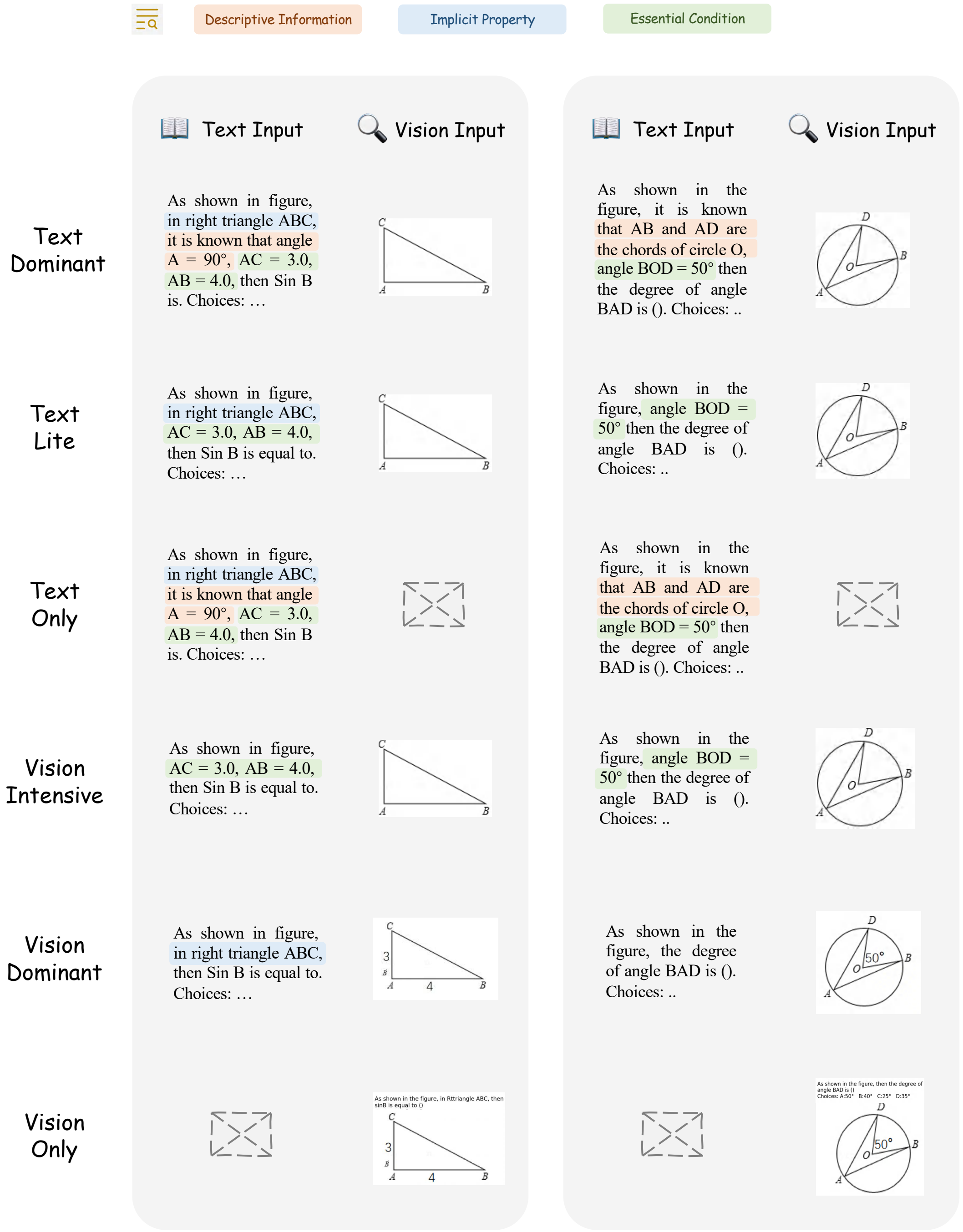

Figure 22: **Comparison of Six Problem Versions in MathVerse.**

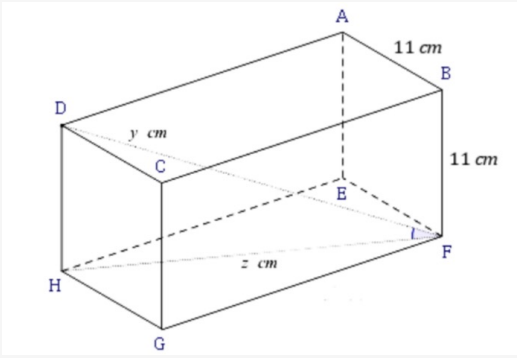

**Question:**

A square prism has a side GF of length 15 cm as shown. Now, we want to find y, the length of the diagonal DF. Calculate y to two decimal places.

**Answer:** y = 21.61

**Subject:** Solid Geometry

**Subfield:** Length

**GPT-4V:**

✓ 1/1 1. Consider the square prism as a combination of two right-angled triangles. Triangle DEF is one of them, and triangle DCF is the other.

✓ 1/1 2. Calculate the length of the diagonal EF of the bottom square face EFGH using the Pythagorean theorem. Since EF is the diagonal of a square of side length 15cm (side GF), we have: EF^2 = GF^2 + GE^2, EF^2 = 450. Now taking the square root to find EF: EF = √450 = 15√2 centimeters.

✗ 0/1 3. Find the diagonal DF which runs through the rectangular face DCEF. Triangle DEF is a right-angled triangle, hence we again apply Pythagorean theorem, with EF (just calculated) and DE being the legs, and DF being the hypotenuse. Since DE is equivalent in length to AB which is given as 11 cm, we have: DF^2 = DE^2 + EF^2 = 571.Taking the square root to find DF: DF = √571. Let's calculate the numeric value to two decimal places: DF ≈ 23.89 centimeters.

✗ 0/1 4. Therefore, y, the length of the diagonal DF, is approximately 23.89 cm.

Visual Perception Error

Multi-step Score: 0.5 Final CoT Evaluation Score: 0.35

**LLaVA-NeXT:**

✓ 1/1 1. Use the Pythagorean theorem to find the length of the diagonal. The theorem states that the square of the length of the hypotenuse (the diagonal in this case) is equal to the sum of the squares of the lengths of the legs. In this case, the legs are the sides of the square prism, which are of equal length.

✗ 0/1 2. Calculate the square of the lengths of the legs: y^2 = 11^2 + 11^2, y^2 = 121 + 121, y^2 = 242.

✗ 0/1 3. Take the square root of both sides to find the length of the diagonal: y = √242.

✗ 0/1 4. Round the square root to two decimal places and then multiply by itself to find the length of y to two decimal places: y = √242 ≈ 15.5, y * y ≈ 25.25.

✗ 0/1 5. The length of the diagonal DF is approximately 15.5 cm.

Numerical Calculation Error

Multi-step Score: 0.2 Final CoT Evaluation Score: 0.14

**SPHINX-MoE:**

✓ 1/1 1.Use the Pythagorean theorem to find the length of the diagonal DF.

✓ 1/1 2.The Pythagorean theorem states that in a right triangle, the square of the length of the hypotenuse (the side opposite the right angle) is equal to the sum of the squares of the lengths of the other two sides.

✗ 0/1 3.In this case, we have a right triangle with sides of length 11cm and 15cm. Let's call the length of the hypotenuse y.

✗ 0/1 4. Using the Pythagorean theorem, we have: y^2 = 11^2 + 15^2, y^2 = 121 + 225, y^2 = 346.

✗ 0/1 5. Taking the square root of both sides, we get: y = √346, y ≈ 18.60.

✗ 0/1 6. Therefore, the length of the diagonal DF is approximately 18.60 cm.

Multi-step Score: 0.33 Final CoT Evaluation Score: 0.23

Figure 23: **Response Comparison of GPT-4V [47], LLaVA-NeXT [38], and SPHINX-MoE [21].** We adopt the Text-lite version of the problem, and highlight the key-step extraction and scoring by the CoT evaluation strategy.

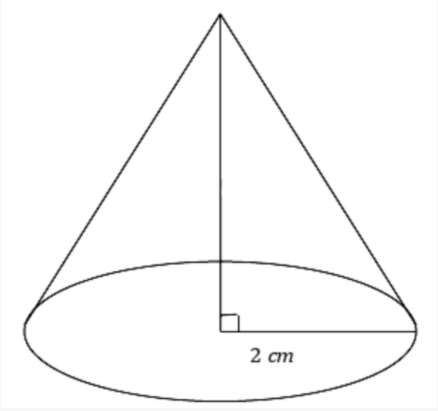

**Question:**

Find the volume of the cone shown. The perpendicular height is 6 cm. Round your answer to two decimal places.

**Answer:** Volume = 25.13 $cm^3$

**Subject:** Solid Geometry

**Subfield:** Volume

**GPT-4V:**

✓ 1/1 1.Use the formula for the volume of a cone: Volume = 1/3 * π * r^2 * h, where r is the radius and h is the height of the cone.

✓ 1/1 2. Substitute the given values into the formula: Volume = 1/3 * π * (2 cm)^2 * 6 cm, Volume = 1/3 * π * 4 $cm^2$ * 6 cm, Volume = π * 4 $cm^2$ * 2 cm, Volume = 8 π $cm^3$.

✓ 1/1 3. Compute the volume numerically using π ≈ 3.14159: Volume ≈ 8 * 3.14159, Volume ≈ 25.13272 $cm^3$.

✓ 1/1 4. Round the result to two decimal places: Volume ≈ 25.13 $cm^3$.

✓ 1/1 5. Final value: 25.13 $cm^3$.

Multi-step Score: 1 Final CoT Evaluation Score: 1

**LLaVA-NeXT:**

Reasoning Error

✓ 1/1 1. Use the formula for the volume of a cone: V = (1/3) π r^2 h, where r is the radius and h is the height of the cone.

✗ 0/1 2. Convert the height from centimeters to meters. There are 100 centimeters in a meter, so 6 centimeters is equal to 0.06 meters.

✗ 0/1 3. Plug the values into the formula: V = (1/3) π (2^2) (0.06), V = (1/3) π (4) (0.06), V = (1/3) π (4) (0.06) $m^3$, V = 0.00064 $m^3$.

✗ 0/1 4. Round to two decimal places. The volume of the cone is approximately 0.00064 cubic meters.

Multi-step Score: 0.25 Final CoT Evaluation Score: 0.175

**SPHINX-MoE:**

Knowledge Error

✓ 1/1 1. Use the formula for the volume of a cone: V = (1/3) π r^2 h, where V is the volume, r is the radius, and h is the height.

✗ 0/1 2. Given the height h = 6 cm but not the radius r. Use the fact that the radius is equal to the height of the triangle formed by the cone's slant height and the height of the cone.

✗ 0/1 3. The slant height is the hypotenuse of a right triangle with legs of length 6 cm and 2 cm (the radius of the base of the cone). Use the Pythagorean theorem to find the slant height: a^2 = b^2 + c^2, where a is the hypotenuse (slant height), b is the height of the cone (6 cm), and c is the radius of the base of the cone.

✗ 0/1 4. Calculate the slant height: a^2 = 6^2 + 2^2, a^2 = 36 + 4, a^2 = 40, a = √40.

✗ 0/1 5. Use the Pythagorean theorem again to find the radius (r = 2 cm): r^2 = a^2 - b^2, r^2 = (√40)^2 - 6^2 = 40 - 36, r^2 = 4, r = √4.

✗ 0/1 6. Now that we have the radius (r = 2 cm), we can find the volume of the cone: V = (1/3) π r^2 h, V = (1/3) π (2)^2 (6), V = (1/3) π (4)(6), V = (1/3) π (24), V = 8π.

✓ 1/1 7. The volume of the cone is approximately 25.13 cubic centimeters.

Multi-step Score: 0.29 Final CoT Evaluation Score: 0.5

Figure 24: **Response Comparison of GPT-4V [47], LLaVA-NeXT [38], and SPHINX-MoE [21].** We adopt the Text-lite version of the problem, and highlight the key-step extraction and scoring by the CoT evaluation strategy.

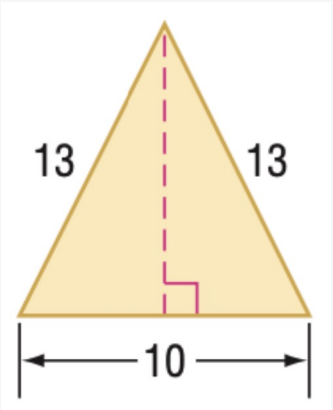

**Question:**

There is an isosceles triangle. Find the area of the figure.
Choices: A:30 B:60 C:120 D:240

**Answer:** B

**Subject:** Plane Geometry

**Subfield:** Area

**GPT-4V:**

✓ 1/1 1. Use the formula for the area of a triangle, which is ( 1/2 * base * height).

✓ 1/1 2. Determine the height h of the isosceles triangle with two equal sides (each with length 13) and a base (length 10).

✓ 1/1 3. Drop a perpendicular from the vertex between the equal sides to the midpoint of the base to create two right-angled triangles. This splits the base into two segments of length 5 each.

✓ 1/1 4. Use the Pythagorean theorem for one of these right-angled triangles: h^2 + 5^2 = 13^2, h^2 = 13^2 - 5^2, h^2 = 169 – 25, h^2 = 144. So, h = √144. Therefore, h = 12.

✓ 1/1 5. Find the area of the isosceles triangle: Area = (1/2 * base * height), Area = (1/2 * 10 * 12), Area = 5 * 12, Area = 60.

✓ 1/1 6. The area of the isosceles triangle is 60. The correct option is B: 60.

Multi-step Score: 1 Final CoT Evaluation Score: 1

**LLaVA-NeXT:**

✓ 1/1 1. Calculate the area of an isosceles triangle using the formula: Area = (1/2) * Base * Height.

✓ 1/1 2. In the given problem, the base of the triangle is 10 units long, and the height is 13 units long.

✗ 0/1 3. Substitute the given values into the formula: Area = (1/2) * 10 * 13, Area = 13 * 7, Area = 91.

✗ 0/1 4. The final answer is 91.

Numerical Calculation Error

Multi-step Score: 0.5 Final CoT Evaluation Score: 0.35

**SPHINX-MoE:**

Visual Perception Error

✓ 1/1 1. The area of an isosceles triangle is given by the formula A = (1/2) * b * h, where b is the base length and h is the height.

✗ 0/1 2. In this case, the base length is 10 and the height is 13.

✗ 0/1 3. Plugging these values into the formula, we get A = (1/2) * 10 * 13 = 65.

✗ 0/1 4. Therefore, the area of the triangle is 65. The correct option is B: 60.

Multi-step Score: 0.25 Final CoT Evaluation Score: 0.475

Figure 25: **Response Comparison of GPT-4V [47], LLaVA-NeXT [38], and SPHINX-MoE [21].** We adopt the Text-lite version of the problem, and highlight the key-step extraction and scoring by the CoT evaluation strategy.

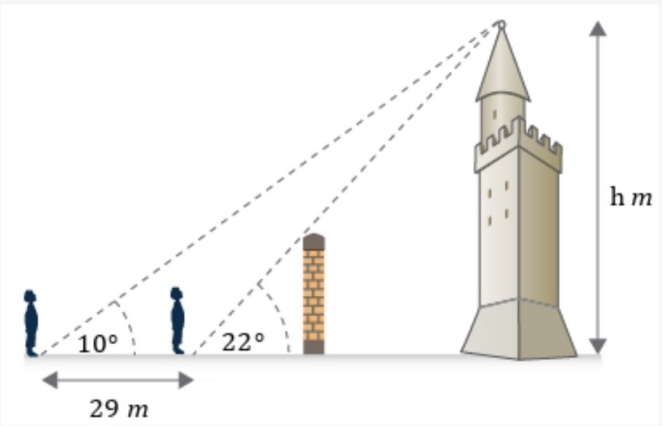

**Question:**

Find the value of h to the nearest metre.

**Answer:** 9

**Subject:** Plane Geometry

**Subfield:** Applied

**GPT-4V:**

✓ 1/1 1. Use trigonometry, specifically the tangent function, to relate the opposite side of the angle in the triangle (which is h, the height of the tower) to the adjacent side (which is the horizontal distance from the base of the tower).

✓ 1/1 2. For the smaller angle of 10 degrees, the tower with the known height creates the opposite side, and the 29-meter distance is the adjacent side. However, we need to find the height of the taller tower, for which we have a 22-degree angle.

✓ 1/1 3. Use the tangent function for the 22-degree angle: tan(22°) = opposite/adjacent tan(22°) = h/29.

✓ 1/1 4. Solve for h: h = 29 * tan(22°).

✓ 1/1 5. Calculate this with a calculator: h ≈ 29 * 0.4040 (tan(22°) ≈ 0.4040) h ≈ 11.716 metres.

✓ 1/1 6. Round to the nearest metre: h ≈ 12 metres Final value: 12.

Multi-step Score: 1 Final CoT Evaluation Score: 1

**LLaVA-NeXT:**

✓ 1/1 1. Determine the height of the tower relative to the person's height by measuring the distance between the person's shadow and the tower.

✗ 0/1 2. Assume that the distance between the person and the tower is equal to the length of the shadow.

✓ 1/1 3. Use the formula to find the length of the shadow: Shadow length = (person's height) x tan(angle of light).

✗ 0/1 4. In this case, the angle of light is approximately 90 degrees, so the shadow length is: Shadow length = 2.2 meters x tan(90 degrees) = 2.2 meters x 1 = 2.2 meters.

Visual Perception Error

✗ 0/1 5. Find the height of the tower relative to the person's height by subtracting the shadow length from the tower's height: h = 2.5 meters - 2.2 meters = 0.3 meters.

✗ 0/1 6. The value of h is approximately 0.3 meters, or 30 centimeters.

Reasoning Error

Multi-step Score: 0.3 Final CoT Evaluation Score: 0.2

**SPHINX-MoE:**

✗ 0/1 1. The single answer is that the value of h is 30.

Multi-step Score: 0 Final CoT Evaluation Score: 0

Figure 26: **Response Comparison of GPT-4V [47], LLaVA-NeXT [38], and SPHINX-MoE [21].** We adopt the Text-lite version of the problem, and highlight the key-step extraction and scoring by the CoT evaluation strategy.

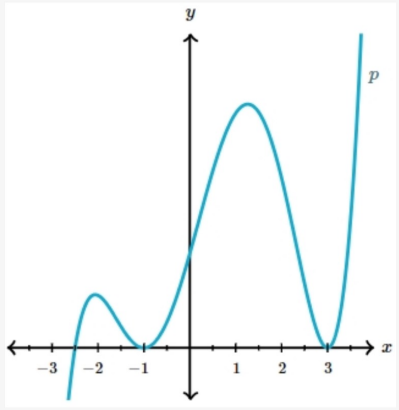

**Question:**

A polynomial p is graphed. What could be the equation of p ?
Choices:
A: $p(x)=(x+1)^2(2x+5)^2(x-3)^2$
B: $p(x)=(x+1)^2(2x+5)(x-3)^2$
C: $p(x)=(x+1)^2(2x+5)(x-3)$
D: $p(x)=(x+1)(2x+5)(x-3)^2$

**Answer:** B **Subject:** Functions **Subfield:** Expression

**GPT-4V:**

✓ 1/1 1. Analyze the graph for key features such as x-intercepts (roots) and the behavior of the graph at those intercepts.

✓ 1/1 2. Identify the x-intercepts: at x = -1, x = -2.5 (which is -5/2 when converted to a fraction), and x = 3. This tells us that the factors of the polynomial must include (x + 1), (x + 5/2), and (x - 3).

✓ 1/1 3. Look at the behavior of the graph at each intercept to determine the multiplicity of each root. The graph touches the x-axis and turns around at x = -1 and x = 3, which indicates that these roots are of even multiplicity. The graph crosses the x-axis at x = -2.5, which indicates that this root is of odd multiplicity.

✓ 1/1 4. Match the graph to the correct equation by comparing the multiplicity of the roots with the given options.

✓ 1/1 5. The correct equation that matches the graph is: B: $p(x)=(x+1)^2(2x+5)(x-3)^2$.

Multi-step Score: 1 Final CoT Evaluation Score: 1

**LLaVA-NeXT:**

Visual Perception Error

✓ 1/1 1. Analyze the graph and identify the degree of the terms in the polynomial.

✗ 0/1 2. Identify that the polynomial has a degree of 2 for the terms with the highest degree, which are the squares of the linear terms.

✗ 0/1 3. Identify the highest degree term that is not squared is the term with the variable raised to the second power, which is $(x+1)^2$.

✗ 0/1 4. Identify the next highest degree term is $(2x+5)^2$, and the lowest degree term is $(x-3)^2$.

✗ 0/1 5. Conclude that the equation of the polynomial p(x) should be of the form: $p(x) = (x+1)^2(2x+5)^2(x-3)^2$ This is option A.

Multi-step Score: 0.2 Final CoT Evaluation Score: 0.14

**SPHINX-MoE:**

✗ 0/1 1. The graph of p has a vertical tangent at x=-1.

✗ 0/1 2.The graph has a horizontal tangent at x=3.

Visual Perception Error

✗ 0/1 3. The graph has a vertical tangent at x=-5/2.

✓ 1/1 4.The graph of p also passes through the point (-1,0).

✗ 0/1 5. The equation of p is $p(x)=(x+1)^2(2 x+5)(x-3)^2$.

✗ 0/1 6. The correct option is A.

Multi-step Score: 0.17 Final CoT Evaluation Score: 0.12

Figure 27: **Response Comparison of GPT-4V [47], LLaVA-NeXT [38], and SPHINX-MoE [21].** We adopt the Text-lite version of the problem, and highlight the key-step extraction and scoring by the CoT evaluation strategy.

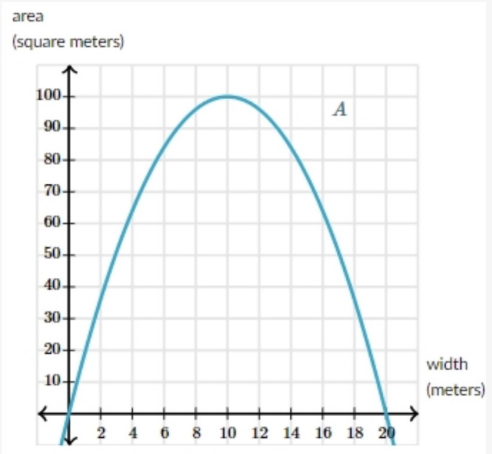

**Question:**

The function A models the rectangle's area (in square meters) as a function of its width (in meters). Which of these statements are true? Choices:

A: Greater width always relates to smaller area.
B: Greater width relates to smaller area as long as the width is less than 10 m.
C: When there is no width, the area is 20 $m^2$.
D: When there is no width, the area is 0 $m^2$.

**Answer:** D **Subject:** Functions **Subfield:** Applied

**GPT-4V:**

✓ 1/1 1.Analyze the graph of the function A, which models the area of a rectangle as a function of its width.

✓ 1/1 2. Evaluate each statement based on the graph:
- A: Greater width always relates to a smaller area.
- B: The area is always increasing with the width.
- C: When the width is 0, the area is not 0.
- D: When there is no width, the area is 0 $m^2$.

✓ 1/1 3. Determine that statement D is true because the graph starts at the origin (0,0), indicating that when there is no width, there is no area, which is consistent with the properties of a rectangle.

✓ 1/1 4. The correct option is: D: When there is no width, the area is 0 $m^2$.

Multi-step Score: 1 Final CoT Evaluation Score: 1

**LLaVA-NeXT:**

Visual Perception Error

✓ 1/1 1. Analyze the given information and the graph of the quadratic function A.

✗ 0/1 2. Observe that the vertex of the quadratic function is at (10, 100), indicating that the function has a minimum point at this point.

✓ 1/1 3. Note that the quadratic function intersects the x-axis at (0,0) and (20,0), which means the area is zero when the width is zero.

✓ 1/1 4. Consider the given statements and determine their correctness based on the observations from the graph.

✓ 1/1 5. Statement A is incorrect as the area does not always decrease with increasing width.

✗ 0/1 6. Statement B is correct as the area decreases with increasing width up to the vertex, which is at 10 meters.

✓ 1/1 7. Statement C is incorrect as the area is zero when the width is zero, not 20 square meters.

✓ 1/1 8. Statement D is correct as the area is zero when the width is zero.

✗ 0/1 9. The correct statement is B: Greater width relates to smaller area as long as the width is less than 10 m.

Multi-step Score: 0.67 Final CoT Evaluation Score: 0.47

**SPHINX-MoE:**

✓ 1/1 1. The single answer is D.

Multi-step Score: 1 Final CoT Evaluation Score: 1

Figure 28: **Response Comparison of GPT-4V [47], LLaVA-NeXT [38], and SPHINX-MoE [21].** We adopt the Text-lite version of the problem, and highlight the key-step extraction and scoring by the CoT evaluation strategy.

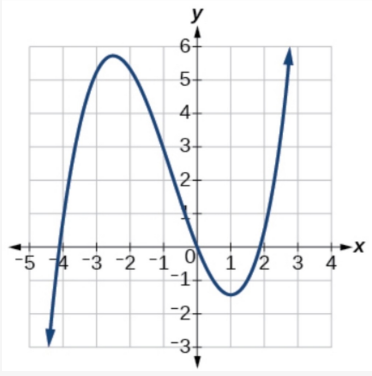

**Question:** (Text Dominant)

There is a curve goes through (-4,1), (-1,3), and (0,0). Estimate the intervals on which the polynomial function is increasing or decreasing.

**Answer:** The function is increasing on (-∞,-2.5) ∪ (1, ∞), and decreasing on (-2.5,1)

**Subject:** Functions

**Subfield:** Property

### Text Dominant:

✓ 1/1 1. Determine the intervals on which the function is increasing or decreasing by looking at the slope of the curve on the graph.

✗ 0/1 2. The function starts off decreasing from the point (-5, something above 6) to the point (-4, 1).

✗ 0/1 3. The function then starts increasing at the point (-4, 1) and continues to increase until it reaches a peak somewhere between -2 and -1 on the x-axis.

✗ 0/1 4. After reaching the peak, the function starts to decrease again, and it continues to decrease past the point (-1, 3) until it reaches the point (0, 0).

✗ 0/1 5. The function reaches its minimum at (0, 0) and then starts to increase again. It continues to increase past the point (2, something below -3) and keeps increasing as it moves off to the right of the graph.

✗ 0/1 6. The intervals on which the function is increasing or decreasing are as follows:
- Decreasing: from the start of the graph to x = -4, and from just before x = -2 to x = 0.
- Increasing: from x = -4 to just before x = -2, and from x = 0 to the end of the graph.

✗ 0/1 7. More precisely:
- The function is increasing on the intervals (-4, -1.5) and (0, ∞).
- The function is decreasing on the intervals (-∞, -4) and (-1.5, 0).

✓ 1/1 8. Note that the exact turning points are estimated since the graph does not provide precise values, and the intervals are given in terms of x-values. The exact values of the turning points cannot be determined without more information or a more precise graph.

Multi-step Score: 0.25 Final CoT Evaluation Score: 0.175

### Text Lite:

✓ 1/1 1. Determine the slope of the function as it moves from left to right along the x-axis.

✓ 1/1 2. Identify the points where the function reaches its peaks and its lowest points.

✗ 0/1 3. The function is increasing from the leftmost part of the graph until it reaches its first peak, which is approximately at x = -3.

✗ 0/1 4. The function starts decreasing after this peak until it reaches its lowest point, which is approximately at x = 0.

✗ 0/1 5. The function then starts increasing again from x = 0 until it reaches its next peak, which is approximately at x = 2.

✗ 0/1 6. The function decreases again after x = 2 towards the rightmost part of the graph.

✗ 0/1 7. The intervals on which the function is increasing are from x = -5 to x = -3 and from x = 0 to x = 2. The intervals on which the function is decreasing are from x = -3 to x = 0 and from x = 2 to x = 4.

✗ 0/1 8. Final values:
Increasing intervals: (-5, -3) and (0, 2),
Decreasing intervals: (-3, 0) and (2, 4).

Multi-step Score: 0.25 Final CoT Evaluation Score: 0.175

### Vision Dominant:

✓ 1/1 1. Determine the intervals on which the function is increasing or decreasing by looking at the slope of the function as it moves from left to right along the x-axis.

✗ 0/1 2. From the graph, identify that the function is increasing from the leftmost part of the graph until it reaches its first peak. This occurs from approximately x = -5 to x = -2.5.

✗ 0/1 3. After reaching the first peak, the function starts to decrease until it reaches the bottom of the first trough. This occurs from approximately x = -2.5 to x = 0.

✗ 0/1 4. The function then starts to increase again from the bottom of the trough until it reaches the second peak. This occurs from approximately x = 0 to x = 1.

✗ 0/1 5. Finally, the function decreases again after the second peak towards the rightmost part of the graph. This occurs from approximately x = 1 to x = 4.

✗ 0/1 6. Conclude that the function is increasing on the intervals: (-5, -2.5) and (0, 1)

✗ 0/1 7. And the function is decreasing on the intervals: (-2.5, 0) and (1, 4)

Multi-step Score: 0.14 Final CoT Evaluation Score: 0.1

Figure 29: **Response of Different Problem Versions by GPT-4V [47].** By comparing the output from three versions, we observe increasing reasoning errors when more textual content is visually depicted in diagrams. We only showcase the Text-dominant version of the problem for brevity.

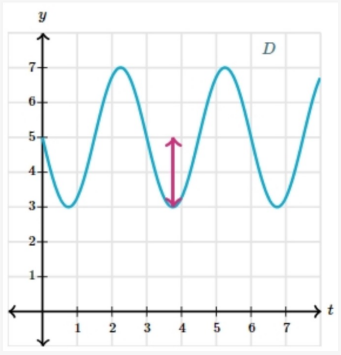

**Question:** (Text Dominant)

Sia is swinging from a chandelier. The horizontal distance between Sia and the wall, in meters, is modeled by D(t) where t is the time in seconds. The function is graphed below, along with one segment highlighted, a red double arrow from (3.75,3), (3.75,5). What is the meaning of the highlighted segment?
Choices:
A:The chandelier is hung 2 meters from the wall.
B:The furthest Sia gets from the point where the chandelier hangs is 2 meters.
C:The chandelier is 2 meters wide.
D:Sia completes 2 swing cycles per second.

**Answer:** B **Subject:** Functions **Subfield:** Applied

Text Dominant:

✓ 1/1 1. The graph shows a function D(t) that represents the horizontal distance between Sia and the wall over time.

✓ 1/1 2. The highlighted segment with a red double arrow spans vertically from the point (3.75, 3) to the point (3.75, 5) on the graph.

✗ 0/1 3. This segment is at a constant time of 3.75 seconds and shows a change in the distance D(t) from 3 meters to 5 meters.

✗ 0/1 4. The segment does not provide information about the width of the chandelier or the number of swing cycles per second. Instead, it indicates a change in distance at a specific moment in time.

✓ 1/1 5. Since the distance changes from 3 to 5 meters, this represents a 2-meter difference in the horizontal distance from the wall during the swing.

✓ 1/1 6. Therefore, the meaning of the highlighted segment is that the furthest Sia gets from the point where the chandelier hangs is 2 meters.

✓ 1/1 7. The correct option is: B: The furthest Sia gets from the point where the chandelier hangs is 2 meters.

Multi-step Score: 0.71 Final CoT Evaluation Score: 0.8

Text Lite:

✓ 1/1
✓ 1/1 1. Analyze the graph which shows a periodic function, representing the back-and-forth motion of Sia swinging from the chandelier. The horizontal axis is labeled with time t in seconds, and the vertical axis is labeled with distance D in meters. The highlighted segment on the graph indicates a change in the distance from one point to another over time.

2. Evaluate each choice:
✓ 1/1 - A: The chandelier is hung 2 meters from the wall. This statement would imply a fixed distance from the wall, which the graph does not provide information about.
✗ 0/1 - B: The furthest Sia gets from the point where the chandelier hangs is 2 meters. This statement seems to be about the amplitude of the swing, but the highlighted segment does not indicate the amplitude; it indicates a change in distance over time.
✓ 1/1 - C: The chandelier is 2 meters wide. The width of the chandelier is not something that can be determined from this graph, which shows distance from the wall over time.
✓ 1/1 - D: Sia completes 2 swing cycles per second. The highlighted segment shows a change in distance over time, but it does not provide information about the number of cycles per second.

✗ 0/1 3. Conclude that none of the options A, B, C, or D directly describe the meaning of the highlighted segment in terms of the graph's representation of distance over time.

✗ 0/1 4. Given the options provided and the information from the graph, none of the choices accurately describe the meaning of the highlighted segment. Therefore, based on the information provided and the reasoning above, there is no correct option to choose from A, B, C, or D.

Multi-step Score: 0.63 Final CoT Evaluation Score: 0.44

Vision Dominant:

✓ 1/1 1. The graph shows a periodic function, representing the motion of a pendulum or a swing. The horizontal axis represents time (t), and the vertical axis represents the horizontal distance (D) from the wall. The highlighted segment indicates a change in the distance D over time.

✗ 0/1 2. The highlighted segment is a vertical line, showing a change in distance at a particular moment in time, from a lower value to a higher value on the D(t) function.

✗ 0/1 3. None of the options A, B, C, or D directly relate to the meaning of a single vertical segment on the graph. The closest interpretation of the highlighted segment would be the change in distance from the wall at a specific moment, which is not explicitly described in any of the options provided.

✗ 0/1 4. Given the information and the graph, none of the options A, B, C, or D correctly describe the meaning of the highlighted segment.

Multi-step Score: 0.25 Final CoT Evaluation Score: 0.175

Figure 30: **Response of Different Problem Versions by GPT-4V [47].** By comparing the output from three versions, we observe increasing reasoning errors when more textual content is visually depicted in diagrams. We only showcase the Text-dominant version of the problem for brevity.

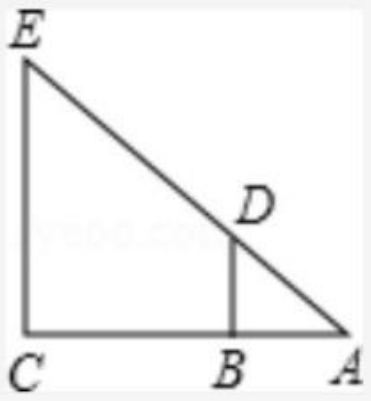

**Question:** (Text Dominant)

As shown in the figure, Xiaoming designed two right angles to measure the width of the river BC, he measured AB = 2.0, BD = 7/3, CE = 9.0, then the width of the river BC is ()
Choices:
A:7 / 2
B:17 / 2
C:40 / 7
D:11

**Answer:** C **Subject:** Plane Geometry **Subfield:** Applied

Text Dominant:

✓ 1/1 1. Identify that the triangles ABD and CEB are similar because they both have a right angle, and they share the angle at point B.

✓ 1/1 2. Use the property of similar triangles that the ratios of their corresponding sides are equal:AB/BD = BC/CE.

✓ 1/1 3. Substitute the given values into the ratio: 2.0 / (7.0/3.0) = BC / 9.0.

✓ 1/1 4. Cross-multiply to solve for BC: 2.0 * 9.0 = BC * (7.0/3.0).

✓ 1/1 5. Solve for BC by dividing both sides by (7.0/3.0): BC = 18.0 / (7.0/3.0).

✓ 1/1 6. Simplify the equation: BC = 18.0 * (3.0/7.0).

✓ 1/1 7. Calculate the final value: BC = 54.0 / 7.0.

✓ 1/1 8. Round the result to the closest answer: BC = 7.71428571...

✓ 1/1 9. The correct option is: C: 40 / 7 meters.

Multi-step Score: 1 Final CoT Evaluation Score: 1

Text Lite:

✓ 1/1 1. Identify that the triangles ABD and EBC are similar because they both have a right angle, and they share the angle at point B.

✓ 1/1 2. Find the ratio between the sides of the triangles ABD and EBC using the given lengths AB and BD. The ratio of AB to BD is 2.0 / (7.0/3.0) = 6.0/7.0.

✓ 1/1 3. Set up a proportion using the sides of the similar triangles: AB/BD = BC/CE.

✓ 1/1 4. Substitute the known values and the ratio into the proportion: (6.0/7.0) = BC/9.0.

✓ 1/1 5. Solve for BC: BC = (6.0/7.0) * 9.0, BC = 54.0/7.0, BC = 7.71428571...

✗ 0/1 6. Compare the calculated value with the provided options. The closest value to this result among the choices given is 40/7 meters, which is approximately 5.71428571 meters. However, none of the options exactly match the calculated value. There might be a mistake in the calculation or in the provided options. Given the options, the closest one is C: 40/7 m.

Multi-step Score: 0.83 Final CoT Evaluation Score: 0.58

Vision Dominant:

✓ 1/1 1. Identify the similar triangles in the figure (Triangle ABD and triangle EBC).

✓ 1/1 2. Note the lengths of the known sides in each triangle.

✓ 1/1 3. Since the triangles are similar, the corresponding sides are proportional. Write the proportions as:

AB/BD = CE/BC.

✗ 0/1 4. Plug in the known values and solve for BC: BC = 9.0 * (7.0/3.0) / 2.0 BC = 10.5.

✗ 0/1 5. The correct answer is not directly listed among the choices. However, since 10.5 meters can be written as 21/2 meters, which is identical to 7/2 times 3 meters, it seems that there might have been a miscalculation in the options or I'm misinterpreting the figure based on the given variables. Please verify the options provided or the figure details to ensure an accurate answer can be given. Assuming the options provided contain a typo and closest to 10.5 (or 21/2) meters is the fraction that also involves 7, the most sensible choice if we round 10.5 to the nearest half is:

A: 7/2 meters (which is 3.5 meters and not the correct answer).

✗ 0/1 6. Given the absence of an accurate option and assuming the calculations are correct, there should be an option that reflects 21/2 meters, which is not present in the given choices. If we are to choose from the closest available option, it would be:

B: 17/2 meters.

✗ 0/1 7. However, 17/2 meters equals 8.5 meters, which is not the result we have calculated. Therefore, there seems to be an error in the options or in the figures provided which needs to be addressed for an accurate selection. Please review the question or the choices for any discrepancies.

Multi-step Score: 0.43 Final CoT Evaluation Score: 0.3

Figure 31: **Response of Different Problem Versions by GPT-4V [47].** By comparing the output from three versions, we observe increasing reasoning errors when more textual content is visually depicted in diagrams. We only showcase the Text-dominant version of the problem for brevity.

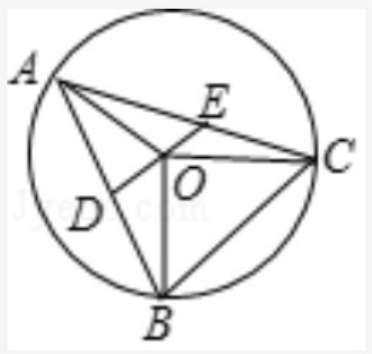

**Question:** (Text Dominant)

As shown in the figure, triangle ABC is inscribed in circle O, OC perpendicular OB, OD perpendicular AB intersects AC at point E. Knowing that the radius of circle O is 1.0, then what is the value of $AE^2 + CE^2$?
Choices:
A:1
B:2
C:3
D:4

**Answer:** B **Subject:** Plane Geometry **Subfield:** Length

Text Dominant:

✓ 1/1 1. Identify that OC is perpendicular to OB, making triangle OCB a right triangle with OC and OB as its legs. Since OC and OB are radii of the circle and the radius is given as 1.0, OC = OB = 1.0. Therefore, triangle OCB is an isosceles right triangle.

✓ 1/1 2. Identify that OD is perpendicular to AB, which means that OD is the altitude of triangle AOB from point O to side AB. Since triangle AOB is also isosceles (OA = OB = 1.0), OD bisects AB. This means that AD = DB.

✓ 1/1 3. Consider triangle AOD and triangle BOD. Since OD is perpendicular to AB, both triangles AOD and BOD are right triangles. Also, since AD = DB, triangle AOD is congruent to triangle BOD. Therefore, AE = EB.

✗ 0/1 4. Since AE = EB and AC = AB (because triangle ABC is inscribed in the circle and AC and AB are both radii), we can say that CE = EB.

✗ 0/1 5. To find AE^2 + CE^2, since AE = CE, we can say that AE^2 + CE^2 = 2 * AE^2.

✗ 0/1 6. To find AE, use the Pythagorean theorem in triangle AOD (or BOD, since they are congruent). We have:

AO^2 = AD^2 + OD^2.

Since AO = 1.0 (radius of the circle) and AD = 1/2 * AB (because AD = DB and AB is the diameter of the circle, which is twice the radius), we have:

1.0^2 = (1/2 * AB)^2 + OD^2.

AB is the diameter of the circle, so AB = 2 * radius = 2 * 1.0 = 2.0. Therefore: 1.0^2 = (1/2 * 2.0)^2 + OD^2 OD^2 = 0
Since OD^2 = 0, OD must be 0, which means that point D coincides with point O, and AE = EB = 1.0 (since AE = EB and AB = 2 * radius = 2.0). Therefore, AE^2 + CE^2 = 2 * AE^2 = 2 * 1.0^2 = 2.

✓ 1/1 7. The correct answer is: B: 2

Multi-step Score: 0.57 Final CoT Evaluation Score: 0.7

Text Lite:

✓ 1/1 1. Use the properties of a circle and the Pythagorean theorem.

✓ 1/1 2. Since OC is perpendicular to OB and both are radii of the circle, triangle OCB is an isosceles right triangle. Therefore, the length of BC can be found using the Pythagorean theorem: BC = √2

✓ 1/1 3. Similarly, since OD is perpendicular to AB, triangle OAD is also an isosceles right triangle. Therefore, AD = OD = 1.

✗ 0/1 4. Using the Pythagorean theorem again, AO = √2. Since AO is the diameter of the circle, AC is also the diameter, and thus AC = AO = √2.

✗ 0/1 5. We have AC = √2 and BC = √2. Since E is a point on AC, AE + EC = AC.

✗ 0/1 6. We want to find AE^2 + CE^2. Let's denote AE as x and CE as (√2 - x). Then we can express AE^2 + CE^2 as: 2x^2 - 2√2x + 2.

✗ 0/1 7. To minimize this expression, we can take the derivative with respect to x and set it to zero to find the minimum value. However, since we are given that AE and CE are segments of a line, and we know that the sum of their squares in a right triangle is constant, we can infer that the sum of the squares of AE and CE will be equal to the square of AC.

✓ 1/1 8. Therefore, AE^2 + CE^2 = AC^2 = (√2)^2 = 2.

✓ 1/1 9. The correct answer is: B: 2

Multi-step Score: 0.56 Final CoT Evaluation Score: 0.69

Vision Dominant:

✓ 1/1 1. Given that the radius of circle O is 1, this means that OA, OE, and OC are all radii of the circle, and thus they all have a length of 1.

✓ 1/1 2. Use the Pythagorean theorem (a^2 + b^2 = c^2) on triangle AOE, where AE is the hypotenuse, and OA and OE are the two legs:

AE^2 = OA^2 + OE^2, AE^2 = 2.

✓ 1/1 3. By the same logic, apply the Pythagorean theorem to triangle COE, where CE is the hypotenuse:

CE^2 = OC^2 + OE^2, CE^2 = 2.

✗ 0/1 4. Now, according to the question, find the value of AE^2 + CE^2:

AE^2 + CE^2 = 2 + 2 AE^2 + CE^2 = 4.

✗ 0/1 5. Therefore, the value of AE^2 + CE^2 is 4. The correct option is D: 4.

Multi-step Score: 0.6 Final CoT Evaluation Score: 0.42

Figure 32: **Response of Different Problem Versions by GPT-4V [47].** By comparing the output from three versions, we observe increasing reasoning errors when more textual content is visually depicted in diagrams. We only showcase the Text-dominant version of the problem for brevity.

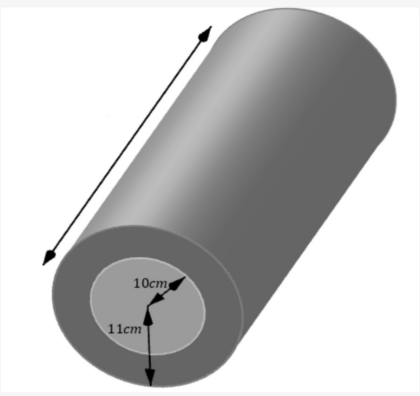

**Question:** (Text Dominant)

Consider the solid pictured and answer the following: Hence what is the total surface area? The outer radius is 11 cm, inner radius is 10 cm and length 24 cm. Give your answer to the nearest two decimal places.

**Answer:** SA = 3298.67 $cm^2$

**Subject:** Solid Geometry

**Subfield:** Area

### Text Dominant

✓ 1/1 1. Calculate the surface area of the outer side of the cylinder using the formula for the lateral surface area of a cylinder: 2π rh, where r is the outer radius and h is the length (height) of the cylinder. Outer surface area = 2 π * 11 * 24.

✓ 1/1 2. Calculate the surface area of the inner side of the cylinder using the same formula, but with the inner radius. Inner surface area = 2 π * 10 * 24.

✓ 1/1 3. Calculate the surface area of both circular ring-shaped ends. This is the area of the outer circle minus the area of the inner circle, multiplied by 2 (since there are two ends). Area for one end = Area of outer circle - Area of inner circle = $\pi r_{outer}$^2 - $\pi r_{inner}$^2 = π11^2 - π10^2. Total area for both ends = 2 * (π(11^2 - 10^2)).

✗ 0/1 4. Add all these together to get the total surface area: Total surface area = 528π + 480π + 42π = 1050π.

✓ 1/1 5. Use the approximate value of π, which is 3.14, to calculate the total surface area in decimal form: Total surface area ≈ 1050 * 3.14.

✗ 0/1 6. Round the result to the nearest two decimal places: 3297.00.

✗ 0/1 7. The total surface area of the solid, to the nearest two decimal places, is approximately 3297.00 square centimeters.

Multi-step Score: 0.57 Final CoT Evaluation Score: 0.4

### Text Lite

✗ 0/1 1. Calculate the area A of a circle using the formula A = πr^2, where r is the radius. In this case, the radius r is 5 cm. Therefore, the area of one circular base is π(5^2) or 25π.

✗ 0/1 2. Since a cylinder has two bases (top and bottom), multiply the area of one base by 2 to find the total area of both bases: 2 × 25π = 50π cm².

✗ 0/1 3. To find the area of the curved surface of the cylinder, use the formula for the lateral surface area of a cylinder, which is A = 2πrh, where h is the height or length of the cylinder. In this case, the height h is 24 cm. The lateral surface area would therefore be 2πrh = 2π(5)(24) = 240π cm².

✗ 0/1 4. Add both calculated areas together to find the total surface area: Total surface area = area of both bases + lateral surface area. Total surface area = 50π + 240π = 290π cm².

✗ 0/1 5. Use the approximation π ≈ 3.14159 to compute the numerical answer: Total surface area ≈ 290 × 3.14159. Total surface area ≈ 910.8611 cm².

✗ 0/1 6. Rounded to the nearest two decimal places, the total surface area is 910.86 cm² .

Multi-step Score: 0 Final CoT Evaluation Score: 0

### Vision Dominant

✗ 0/1 1. Find the area of one circular base using the formula A = πr², where r is the radius of the base. In this case, the radius is 5 cm, so the area of one base is π * 25 cm² = 78.54 cm² (to two decimal places).

✗ 0/1 2. Since there are two circular bases, double the result to get the total area of the bases: 2 * 78.54 cm² = 157.08 cm².

✗ 0/1 3. Calculate the rectangular surface area, which is the curved surface that wraps around the cylinder. The length of the rectangle is the same as the circumference of the base circle, which can be found with the formula C = 2πr. In this case, the circumference is 2 * π * 5 cm = 31.42 cm (to two decimal places).

✗ 0/1 4. The width of the rectangle is the same as the height (length) of the cylinder, which is given as 24 cm. So the area of the rectangle is 31.42 cm * 24 cm = 753.96 cm² (to two decimal places).

✗ 0/1 5. Add the areas of both circles and the rectangular wrap to get the total surface area: 157.08 cm² + 753.96 cm² = 911.04 cm².

✗ 0/1 6. The total surface area of the cylinder to the nearest two decimal places is 911.04 cm² .

Multi-step Score: 0 Final CoT Evaluation Score: 0

Figure 33: **Response of Different Problem Versions by GPT-4V [47].** By comparing the output from three versions, we observe increasing reasoning errors when more textual content is visually depicted in diagrams. We only showcase the Text-dominant version of the problem for brevity.

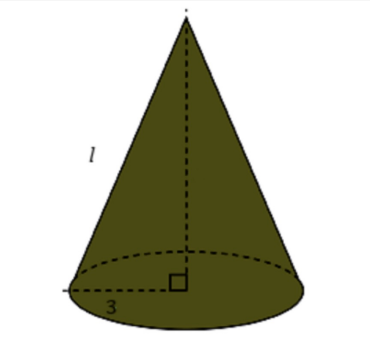

**Question:** (Text Dominant)

Find the volume of the cone pictured. The perpendicular height is 6 and radius is 3. (Give your answer correct to 2 decimal places.)

**Answer:** 56.55

**Subject:** Solid Geometry

**Subfield:** Volume

### Text Dominant

✓ 1/1 1. Use the formula for the volume of a cone: V = 1/3πr^2h , where V is the volume, r is the radius of the base, and h is the height of the cone.

✓ 1/1 2. Given the radius r = 3 and the height h = 6, plug these values into the formula: V = 1/3π(3)^2 (6) , V = 1/3 π ∗ 9 * 6 , V = 18 π.

✓ 1/1 3. Calculate the numerical value using the approximation π ≈ 3.14159: V ≈ 18 * 3.14159 , V ≈ 56.54862 π.

✓ 1/1 4. Round to two decimal places, the volume of the cone is approximately: V ≈ 56.55.

✓ 1/1 5. Final Value: 56.55.

Multi-step Score: 1    Final CoT Evaluation Score: 1

### Text Lite

✓ 1/1 1. Use the formula for the volume of a cone: Volume = 1/3 π r^2 h, where r is the radius of the base of the cone, and h is the height.

✓ 1/1 2. Assume the "3" refers to the radius of the base of the cone, and the perpendicular height provided in the question text is "6". These values are the radius and height respectively.

✓ 1/1 3. Substitute r = 3 and h = 6 into the formula: Volume = 1/3 π (3)^2 (6), Volume = 1/3 π (9) (6), Volume = 1/3 π (54), Volume = 18 π.

✓ 1/1 4. Calculate the numeric value using the approximation π ≈ 3.14159: Volume = 18 * 3.14159, Volume ≈ 56.54862.

✓ 1/1 5. Round to two decimal places, the volume of the cone is: Volume ≈ 56.55.

✓ 1/1 6. Final value: 56.55.

Multi-step Score: 1    Final CoT Evaluation Score: 1

### Vision Dominant

✓ 1/1 1. Identify the shape of the solid figure. In this case, it is a cone.

✓ 1/1 2. Use the formula for the volume of a cone: V = 1/3πr^2 h, where V is the volume, r is the radius of the base, h is the height of the cone, and \pi is the constant approximately equal to 3.14159.

✗ 0/1 3. Identify the values for r and h from the image. Here, h = 6 and the diameter of the base is 3, which means the radius r = 3/2 = 1.5.

✗ 0/1 4. Substitute the values into the formula: V = 1/3π(1.5)^2 * 6, V = 1/3π* 2.25 * 6, V = 1/3π* 13.5, V = π * 4.5, V ≈ 3.14159 * 4.5, V ≈ 14.13765.

✓ 1/1 5. Round the final answer to two decimal places. The volume is 14.14 cubic units.

✗ 0/1 6. Final answer: 14.14.

Multi-step Score: 0.5    Final CoT Evaluation Score: 0.35

Figure 34: **Response of Different Problem Versions by GPT-4V [47].** By comparing the output from three versions, we observe increasing reasoning errors when more textual content is visually depicted in diagrams. We only showcase the Text-dominant version of the problem for brevity.